\documentclass[journal]{IEEEtran}
\usepackage{algorithm}
\usepackage{algorithmic}
\usepackage{multirow}
\usepackage{lineno,hyperref}
\usepackage{xcolor}
\usepackage{graphicx}
\usepackage{amsmath}
\usepackage{booktabs,comment,balance}
\usepackage{threeparttable, tablefootnote}
\usepackage{array, makecell} 

\begin{document}
%
\title{
 Enhancing 3D Human Pose Estimation Amidst Severe Occlusion with Dual Transformer Fusion 
}
%
%
%

 \author{Mehwish~Ghafoor,
         Arif~Mahmood, Muhammad Bilal \\ 
         \vspace{5mm}\textcolor{blue}{Accepted for publication in IEEE Transactions on Multimedia (TMM)}
 \thanks{M Ghafoor and A Mahmood are with the Center for Artificial Intelligence and Robot Vision (CAI\&RV), Department of Computer Science, Information Technology University, Lahore, Pakistan, emails: \{mehwish.ghafoor, arif.mahmood\}@itu.edu.pk}
 \thanks{M Bilal
 is with Center of Excellence in Intelligent Engineering Systems, and 
Department of Electrical and Computer Engineering, Faculty of Engineering, King Abdulaziz University, Jeddah 21589, Saudi Arabia. Email: meftekar@kau.edu.sa
}
\thanks{Manuscript received December 20, 2023}}

\markboth{IEEE TRANSACTIONS ON MULTIMEDIA, Dec~2023}%
{M Ghafoor et al.: Enhancing 3D Human Pose Estimation Amidst Severe Occlusion with Dual Transformer Fusion }

\maketitle

\begin{abstract}

In the field of 3D Human Pose Estimation from monocular videos, the presence of diverse occlusion types presents a formidable challenge. Prior research has made progress by harnessing spatial and temporal cues to infer 3D poses from 2D joint observations. This paper introduces a Dual Transformer Fusion (DTF) algorithm, a novel approach to obtain a holistic 3D pose estimation, even in the presence of severe occlusions. Confronting the issue of occlusion-induced missing joint data, we propose a temporal interpolation-based occlusion guidance mechanism.
To enable precise 3D Human Pose Estimation, our approach leverages the innovative DTF architecture, which first generates a pair of intermediate views. Each intermediate-view undergoes spatial refinement through a self-refinement schema. Subsequently, these  intermediate-views are fused to yield the final 3D human pose estimation. The entire system is end-to-end trainable. Through extensive experiments conducted on the Human3.6M and MPI-INF-3DHP datasets, our method's performance is rigorously evaluated. Notably, our approach outperforms existing state-of-the-art methods on both datasets, yielding substantial improvements. The code is available here: https://github.com/MehwishG/DTF. 

\end{abstract}


\IEEEpeerreviewmaketitle
\section{Introduction}
\label{sec:introduction}
\IEEEPARstart{3D}{human} pose estimation (HPE) has vast applications in the domain of anomaly detection, action recognition, person re-identification, and augmented reality~\cite{bao2022fusepose}. A lot of work has been done on 3D human pose estimation using images or videos as input. Often 3D human pose estimation has been performed in two stages. First, 2D poses are detected from images and then 2D-to-3D lifting is performed to estimate 3D joint positions. Due to 2D pose detection errors and occlusions, incomplete and/or incorrect 2D poses may be obtained. Such errors may get transferred to the 2D to 3D lifting process resulting in ambiguous 3D poses. To handle such issues, robust 2D to 3D lifting algorithms are required, which is the focus of the current work.

Many recent research works have addressed the 2D-to-3D pose lifting problem in the presence of occlusions or missing joint positions by incorporating the spatio-temporal information present in a sequence of 2D poses extracted from a video ~\cite{cai2019exploiting,pavllo20193d,chen2021anatomy,liu2020attention}. Zhao et al. ~\cite{zhao2019semantic} proposed the graph convolutional network (GCN) for the estimation of 3D poses using 2D joint positions. Zhang et al. ~\cite{zhang2022mixste} employed a spatio-temporal encoder for lifting 2D-to-3D poses. Some existing methods rely on dilated convolutional networks to ensure temporal connectivity between consecutive 2D poses in a sequence for effectively handling occlusions or missing joints~\cite{pavllo20193d,liu2020attention,ghafoor2022quantification}. Most methods rely on temporal dilation or predefined adjacency matrices for temporal relationship modeling. The temporal connectivity in these models remains limited, resulting in performance degradation in the presence of occlusions.  

Transformer-based techniques have shown improved performance in many CV tasks including object detection, and image recognition. More recently, transformer-based approaches have also been employed for 2D-to-3D pose lifting \cite{zheng20213d,shan2022p,li2022mhformer,tang20233d}. These approaches have gained improved performance due to the self-attention mechanism which may capture long-range dependencies within 2D temporal pose sequences. 

Most 2D-to-3D pose lifting methods generate a single solution, however being an inverse problem multiple feasible solutions may exist due to its ill-posed nature. Some 2D-to-3D pose lifting methods aim to generate multiple  solutions from a single 2D pose \cite{khirodkar2021multi,wehrbein2021probabilistic,li2022mhformer}. Some of these methods~\cite{khirodkar2021multi,wehrbein2021probabilistic} do not consider the fusion of information between generated multiple 3D poses to improve pose uplifting  except Li et al. \cite{li2022mhformer}. None of these methods handle severe occlusions in the input 2D pose sequence. 

Estimation of 3D pose from severely occluded 2D joint positions is a challenging task. Existing SOTA methods~\cite{cheng2019occlusion,cheng20203d,shan2022p,einfalt2023uplift,zhang2020object,gu2021exploring} have not extensively explored severely occluded 2D joints as inputs, with the exception of the work by Ghafoor and Mahmood~\cite{ghafoor2022quantification}. In contrast to their work, our proposed DTF approach offers an efficient end-to-end solution for estimating 3D joint positions. It incorporates spatial and temporal information more effectively using a novel occlusion guidance mechanism, intermediate view generation, and improved attention mechanism as compared to the previous approaches.

In the current work, we aim to estimate a complete 3D human pose from severely occluded 2D joint positions. For this purpose, we propose a novel Dual Transformer Fusion (DTF) architecture to first generate a pair of intermediate views which are spatially refined using a self-refinement mechanism. The refined views are then fused together using an information fusion scheme to generate an improved 3D human pose. The whole DTF architecture is trained in an end-to-end fashion minimizing regression loss. In order to handle occlusions in the input 2D poses, we propose an occlusion guidance mechanism that  estimates the missing joint positions using temporal interpolation and a confidence score is assigned based on the time gap. A larger time gap may have larger position error and therefore reduced confidence score. The proposed DTF architecture estimates complete 3D poses including those joints which were missing in the input 2D poses with improved position accuracy.
Our contributions are summarized below:
\begin{itemize}

   \item We propose a novel occlusion guidance mechanism to enhance 3D human pose estimation accuracy in the presence of severe occlusion.  The proposed occlusion guidance mechanism is generic and can be integrated with any 2D to 3D pose lifting method.

   \item We propose a novel Dual Transformer Fusion (DTF) algorithm, an innovative approach tailored to address the complex challenge of estimating  complete 3D human poses from a sequence of occluded 2D poses characterized by sparse joint positions. A pair of high dimensional intermediate views are generated, which are succeeded by a refinement process. These refined views are subsequently fused through an information fusion algorithm, resulting in a notably improved 3D human pose estimation.

  \item Extensive experiments are performed using two datasets including Human3.6M and MPI-INF-3DHP to demonstrate the effectiveness of the proposed algorithm. The proposed DTF algorithm achieves state-of-the-art performance in the presence of severe occlusions.
\end{itemize}

The rest of the paper is organized as follows. In Section \ref{sec:back}  recent approaches for 3D HPE and existing occlusion handling schemes are discussed. In Section \ref{sec:method}  the proposed DTF architecture along with the occlusion guidance mechanism is presented. In Section \ref{sec:exp} the experiments and results are reported and finally, the work is concluded in Section \ref{sec:concluions}. 

\vspace{-3mm}
\section{Related work} 
\label{sec:back}
Existing 3D human pose estimation from monocular images can be performed using two approaches. The first approach is a two-stage method in which 2D poses are detected from an image using 2D pose detectors, and then 2D to 3D lifting is performed using various approaches ~\cite{zheng20213d,ghafoor2022quantification,zhang2022mixste,shan2022p,tang20233d,gong2023diffpose}. The second approach is a single-stage method, the 3D pose is predicted directly from an input image ~\cite{nie2019single,jin2022single,chen2023multi}. Our work is focused on a two-stage approach.

\subsection{3D Pose Estimation Paradigms}

Existing 3D lifting networks are capable to regress 3D poses efficiently using detected 2D poses. Martinez \textit{et al.}~\cite{martinez2017simple} proposed a fully connected residual network to estimate 3D joints using 2D joints. 
Hossain \textit{et al.}~\cite{hossain2018exploiting} applied Long Short Term Memory (LSTM) network to exploit temporal information for 3D pose estimation. Sun \textit{et al.} used bidirectional temporal features utilizing Bi-LSTM as a motion discriminator~\cite{sun2023bidirectional} for 3D pose estimation from videos. Pavlo \textit{et al.}~\cite{cai2019exploiting} regressed 3D pose from videos using Temporal-dilated Convolutional Network (TCN) to capture global contextual information. Their work is extended by Liu \textit{et al.}~\cite{liu2020attention} using the attention mechanism. Graph Convolutional Network (GCN) based approaches have also been proposed to estimate 3D pose by modeling dependency between closely related joints using spatio-temporal graphs ~\cite{zhao2019semantic,cai2019exploiting,hu2021conditional,hua2022weakly}.

TCN and GCN based approaches are effective in modeling short-term dependencies by demonstrating better results on the short input sequences. Recently  transformer-based architectures  have also been used to perform 2D to 3D pose estimation. Transformers can model long-term dependencies by capturing complex relationships between different parts of the input sequence using a self-attention mechanism.

Zheng \textit{et al.}~\cite{zheng20213d} proposed a transformer-based architecture  capturing intra-frame spatial correlation  and inter-frame temporal correlation between joints  with the self-attention mechanism. Li \textit{et al.}~\cite{li2022exploiting} introduced a strided transformer for 3D HPE with reduced computational cost. Zhang \textit{et al.}~\cite{zhang2022mixste} proposed a variant of the basic transformer-based architecture by applying separate spatial and temporal transformer blocks. Li \textit{et al.}~\cite{li2022mhformer} introduced a multi-hypothesis transformer using spatial and temporal information. They initially performed one-to-many mapping followed by many-to-one mapping to infer a 3D pose against the central frame. Tang \textit{et al.}~\cite{tang20233d} introduced spatio-temporal criss-cross attention block. These blocks, when stacked together, incorporate structure-enhanced positional embedding to amplify their functionality. Qian \textit{et al.}~\cite{qian2023hstformer} proposed a hierarchical spatio-temporal transformer, aiming to model multiple levels of joints spatio-temporal correlations in a structured manner. Zhao \textit{et al.}~\cite{zhao2023poseformerv2} proposed an improved version of PoseFormer~\cite{zheng20213d} by leveraging the length of the input sequence in the frequency domain to enhance robustness to noisy 2D joint detection. Wang \textit{et al.}~\cite{wang2023global} used a transformer-based encoder for global context and CNN for local features, while Kang \textit{et al.}~\cite{kang2024diffusion} proposed a diffusion-based model for 3D pose estimation.

These works have shown good performance however their performance may degrade due to occluded 2D poses. The detected 2D poses may remain incomplete due to occlusion by self or others, light intensity variations, and low resolution of the object. Due to incomplete 2D poses, estimated 3D poses may not remain accurate and reliable. In order to deal with this limitation occlusion-aware training is required.

\begin{figure*}[t]
\includegraphics[width=\textwidth,keepaspectratio]{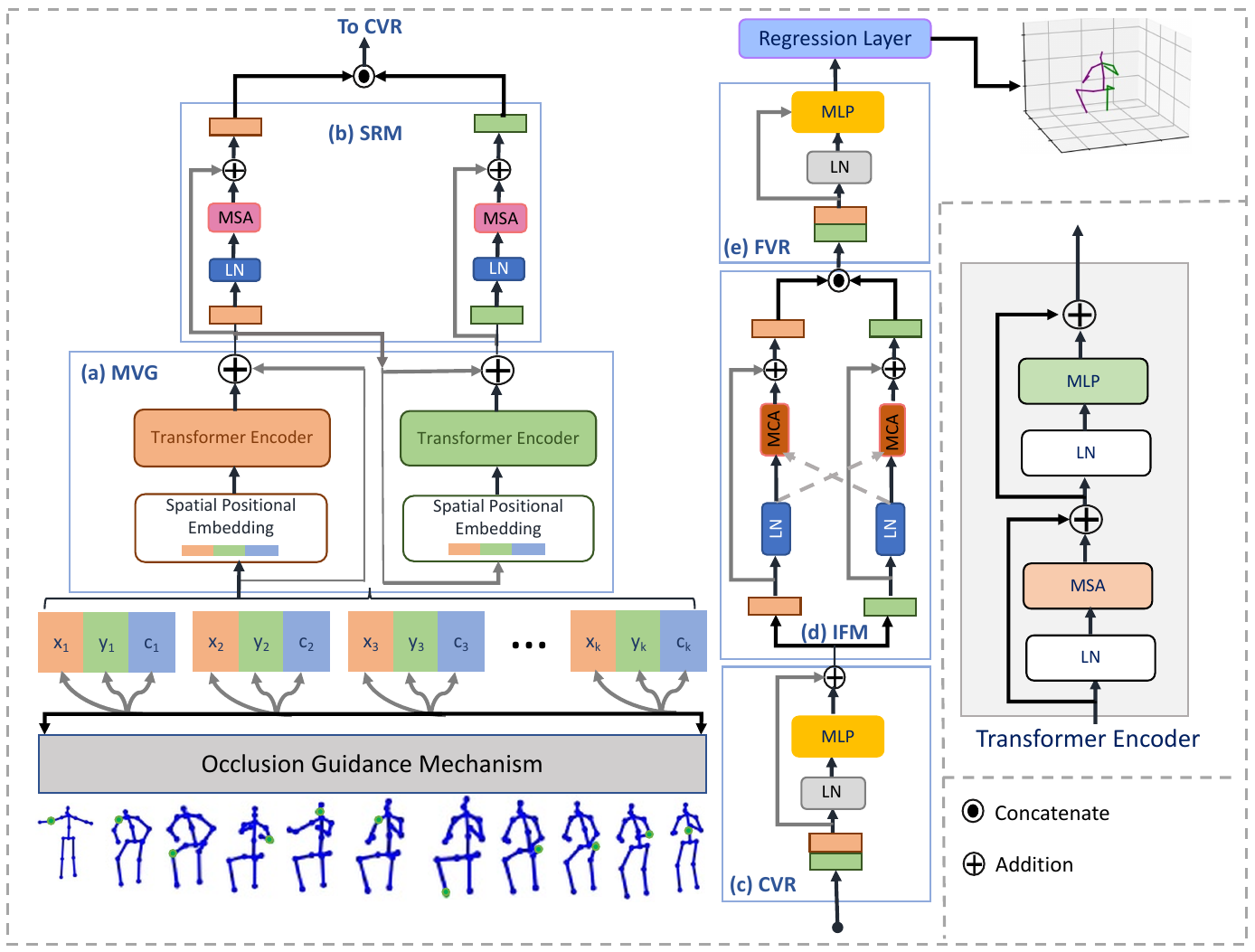}
\caption{System diagram of the proposed  DTF Algorithm: (a) Multi-View Generator (MVG), (b) Self-Refinement Module (SRM) (c) Combined Views Refinement (CVR), (d) Information Fusion Module (IFM), (e) Fused Views Refinement (FVR). Green markers show the available joints in the input sequence.}
\label{fig:arc}
\end{figure*}
\subsection{Occlusion Handling in 3D Pose Estimation}
Some recent works on 3D HPE have proposed occlusion-aware training to generate 3D poses from partially or fully occluded 2D poses \cite{lin2022occlusionfusion}. For instance, Cheng \textit{ et al.}~\cite{cheng2019occlusion} proposed an occlusion-aware network for 3D pose estimation using a cylinder man model. Cheng \textit{et al.}~\cite{cheng20203d} used data augmentation during training to handle occlusion and spatio-temporal networks for 3D pose estimation.

Shan \textit{et al.}~\cite{shan2022p} introduced a two-stage model; In the first stage, sparse input 2D joints are randomly masked spatially and temporally and a complete 2D pose sequence is reconstructed which is fed in the second stage to estimate 3D pose using the many-to-one mapping module that aggregates information from multiple frames. Einfalt \textit{et al.}~\cite{einfalt2023uplift} proposed a two-stage transformer-based method that takes sparse 2D pose sequence and upsampled it using position-aware tokens. The upsampled 2D pose sequence is then passed to generate 3D pose sequence with reduced computational complexity. 
Most of the existing occlusion-aware methods ~\cite{cheng2019occlusion,cheng20203d,shan2022p,einfalt2023uplift,zhang2020object,gu2021exploring} have not quantified their performance using partial and complete occlusion cases. Recently Ghafoor and Mahmood~\cite{ghafoor2022quantification} introduced an occlusion-aware method using indicator variables showing the presence or absence of joints in occluded 2D pose sequence and performed 3D pose estimation using the temporal dilated convolutional network dubbed as T3D-CNN. In contrast, in the current work, a quite different occlusion guidance mechanism is proposed. We generate intermediate views and integrate spatial and temporal information  to generate realistic 3D pose while such improvements have not been utilized by the existing methods \cite{ghafoor2022quantification}.

In contrast to the existing methods~\cite{shan2022p,einfalt2023uplift,ghafoor2022quantification,cheng20203d,cheng2019occlusion}, in the current work, we propose temporal interpolation to fill in the missing joint positions with confidence showing the temporal distance between the current frame and the frame with the available joint. A long-term self-attention mechanism is employed using transformer-based architecture to achieve improved performance. The DTF has shown excellent performance compared to the existing SOTA methods.


\section{Dual Transformer Fusion (DTF) Algorithm }
\label{sec:method}
The system diagram of the proposed DTF algorithm is shown in Fig. \ref{fig:arc}.
Following many previous studies~\cite{pavllo20193d,liu2020attention}, 2D poses are extracted from each video frame in a sequence using an existing 2D pose detector~\cite{chen2018cascaded}. 
The extracted 2D pose sequence is provided as input to the proposed network to estimate a 3D pose corresponding to the center of the input sequence. 
In order to deal with severely occluded 2D pose sequences, we incorporate an occlusion guidance mechanism within our proposed architecture to estimate complete 3D pose with improved position accuracy.
In the following sections, we discuss both the occlusion guidance mechanism and the proposed DTF architecture.

\subsection{Occlusion Guidance Mechanism}
\label{subsec:occ}
Most existing 2D pose detectors provide a confidence score along with each detected 2D joint position. The detection of 2D joint positions may get missed due to several reasons including self-occlusion, occlusion by other objects in the scene, low light, shadows, person truncation, low resolution, and limitations of the underlying 2D pose detectors. Due to one or more of these reasons, many joint positions and associated confidence scores may get missed in the 2D pose sequences.

To handle missing joint information, we propose an occlusion guidance mechanism. We first employ a temporal interpolation scheme to fill in the missing joint positions using the nearest available value of a particular joint. Since we assume a sequence of 2D poses is available, therefore the nearest available joint will be searched in both the past and the future frames. For a missing joint, the closest available joint in terms of the minimum number of frames is selected. Such an interpolation would only provide an approximate position of the missing joint. The position error is corrected by estimating multiple feasible views using DTF architecture.

Assuming in each 2D pose, only $x$ random joints are available such that $x<<j$, where $j$ is the total number of joints. For each missing joint position, the next available joint is searched in both temporal directions. The joints which are available in the current frame are assigned a confidence score of 1.00 while for the joints which are copied from temporal neighbors, the confidence is reduced according to the time gap from the current frame.  We define a search window of size $f_p$ past frames and $f_f$ future frames. The confidence score is reduced by $1/f_p$ for each past neighbor and by $1/f_f$ for each future neighbor. Thus, a reduced confidence score will indicate more error due to increased temporal interval. The confidence score tells the transformer about possible position errors in a particular joint position.

\begin{figure}[t]

\includegraphics[width=7cm]{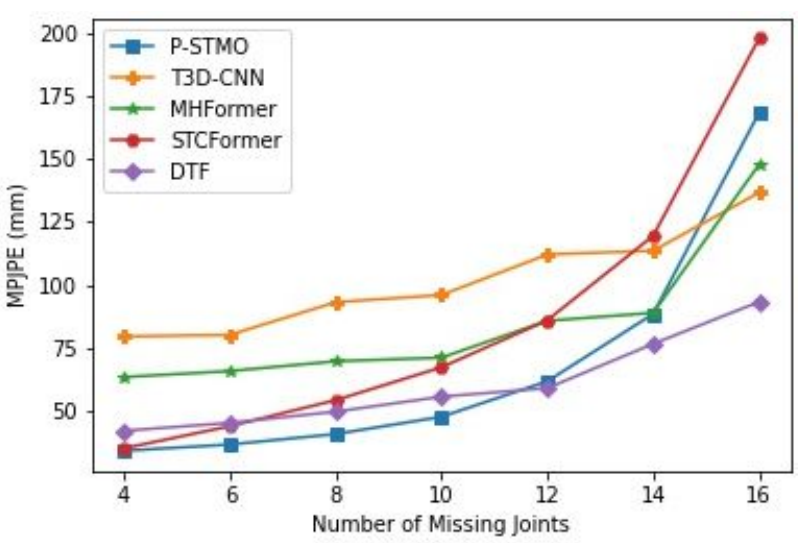}
\caption{Comparison of existing SOTA Methods including T3D-CNN~\cite{ghafoor2022quantification}, P-STMO \cite{shan2022p}, MHFormer \cite{li2022mhformer}, and STCFormer \cite{tang20233d} with proposed DTF algorithm with a varying number of missing joints on MPI-INF-3DHP dataset. As the number of missing joints increases the proposed DTF algorithm has shown a minimum increase in error.}
\label{fig:mpjpe}
\end{figure}
\subsection{Proposed DTF Architecture}
We adopt a baseline transformer architecture that includes multi-head self-attention (MSA) with multilayer perceptron (MLP) \cite{vaswani2017attention}.
MSA is a main block in building a transformer that can be defined as a linear mapping function that takes Query $Q$, Key $K$, and Value $V$ of dimension $d_v$ and outputs the attention matrix. The whole process of self-attention also called scaled dot-product is defined as:
\begin{equation}
    S\mbox{-}Att=Softmax(\frac{QK^\top}{\sqrt d_v})V
\end{equation}
There are $h$ $S\mbox{-}Att$ blocks executing in parallel to represent information at different positions. Later, these $S\mbox{-}Att$ blocks are concatenated to form an output attention matrix.
\begin{equation}
    Z=MSA(Q,K,V)=\texttt{concat}(S\mbox{-}{Att_1},...,S\mbox{-}{Att_h})
\end{equation}

The proposed architecture splits into spatial and temporal domains followed by a regression layer to estimate the complete 3D pose as shown in Fig. \ref{fig:arc}. 
In the spatial domain, a Multi-View Generator is introduced to mimic joint positions in latent space to learn spatial position embedding using a sequence of occlusion-guided 2D joint positions as input to generate multiple intermediate views, which are later passed to MSA and MLP followed by Layer Normalization (LN). The MLP consists of two fully connected layers followed by a non-linear activation function GeLU.

\textbf{Multi-View Generator (MVG):} Let $P \in R^ {t \times j \times 3}$ be an occlusion-guided input 2D pose sequence where $t$ indicates the number of input frames and $j$ is the number of joints and $3$ includes (x,y) coordinates and confidence score. We have used learnable spatial position embedding $E_{sp}^k \in R^ {t \times j \times 3}$ to keep spatial information of 2D poses. $P^k$ is concatenated with spatial position embedding and passed to MVG to generate two views $(k \in [k_1,k_2] )$ as elaborated in the following equations step by step: 
\begin{equation}
    P^k_0 = LN(P^{k})+ E_{sp}^{k}, 
\end{equation}
\begin{equation}
    {\tilde{P}}^k_l = {P}^k_{l-1} + MSA^k (LN({P^k_{l-1}})),
\end{equation}
\begin{equation}
   \widehat{P}^k_l = {\tilde{P}}^k_l + MLP^k(LN({\tilde{P}}^k_l)),
\end{equation}
\begin{equation}
    P_{s}^k = P^{k} + LN(\widehat{{P}}_s^k)).
\end{equation}
We have passed $P^k$ to MSA followed by two MLP layers and layer normalization (LN) to generate $P_s^k$ where $l \in [1,...,s]$ is the index for MVG layers also explained in Fig. \ref{fig:arc} (a). $P_s^k$ are  generated two views with various contextual information which are further refined in the temporal domain.
To ensure temporal consistency between frames, we have exploited temporal information. The generated $k$ views $P_{s}^k$ against each frame are mapped to a high-dimension feature $\hat{Y}^k \in R^{N\times M}$ where $M$ is an embedded dimension. To keep positional information of frames; temporal position embedding $E_{tp}^k \in R^{N \times M} $ along with ${\hat{Y}^k}$ is concatenated to form ${\hat{Y}_0}^ k  = {\hat{Y}^k} + E_{tp}^k $. The intermediate views generated by MVG represent same potential 3D pose, and may differ in depth and the viewing angles. Due to the ambiguous nature of projecting 2D poses to 3D structure, more than one intermediate views are useful. Each view contributes a different configuration that accommodates the 2D evidence yet leads to more accurate possibility of the 3D structure.
The MVG is followed by two modules including the self-refinement and information fusion module.

\noindent \textbf{Self-Refinement Module} (SRM) consists of two Multihead Self-Attention blocks (MSA) and the combined views refinement block (CVR) as represented in Fig. \ref{fig:arc} (b) and (c). MSA block is used for communication within view and is executed in parallel for two views independently as shown in Eq. \ref{eq:f_sa}. CVR block is used to exchange information between two generated views and to refine the representation of views: 
\begin{equation}
\label{eq:f_sa}
    {\widehat{Y}_l}^ k = {\hat{Y}_{l-1}}^ k + MSA(LN({\hat{Y}_{l-1}}^ k)),
\end{equation}
\begin{equation}
\label{eq:mixing}
    {\widehat{Y}}^ {K} = \texttt{concat}(  {\widehat{Y}}^ {k_1},  {\widehat{Y}}^ {k_2} ), 
\end{equation}
\begin{equation}
\label{eq:mixing1}\texttt{concat}(  {\widehat{Y}}^ {k_1} ,  {\widehat{Y}_l}^ {k_2} ) = {\widehat{Y}_l}^ {K} + MLP(LN ( {\widehat{Y}_l}^ {K} )).
\end{equation}
The CVR block has the same configuration as an MLP.

\noindent \textbf{Information Fusion Module} (IFM) consists of cross-view attention (MCA) followed by fused Views refinement module (FVR) as shown in Fig. \ref{fig:arc} (d) and (e). MCA consists of parallel multi-head cross-view attention blocks to measure correlation across views. In order to limit the number of blocks for cross-view attention, we use fixed input of size 2 views only and formulate them as:
\begin{equation}
\label{eq:f-ca}
    {\widehat{Y}_l}^ k = {\widehat{Y}_{l-1}}^ k + MCA^k (LN ( {\widehat{Y}_{l-1}}^ {k_1}), LN({\widehat{Y}_l}^ {k_2}))
\end{equation}
where $k_1$ \& $k_2$ are two generated views.
$FVR$ block has the same functionality as MLP, the output of Eq. \ref{eq:f-ca} is passed to $FVR$ that aggregates all features and generates a single output feature $Y^{K}$ which is fed to the regression layer to generate the 3D pose sequence and the central frame $\tilde{P} \in R^{j \times 3}$ is selected as the final 3D estimation.

The proposed model is trained end to end using the mean per joint position error (MPJPE) as a loss function to minimize the error between the estimated 3D and ground truth joints in each frame.
\begin{equation}
    L_{MPJPE} = \sum_{i=1}^{N} \sum_{j=1}^{J} \lVert  T_{j}^i - \tilde{P}_{j}^i \rVert ^2
\end{equation}

\begin{table*}[!ht]
 \caption{Comparison  of the proposed DTF  with existing SOTA methods  in terms of MPJPE, for $16$  missing joints per frame in Human3.6M dataset using 2D CPN detections as input. Action-wise performance is reported for each algorithm: Directions (Dir.), Discussion (Disc), Eat, Greet, Phoning (Phone), Photo, Pose, Purchases(Purch), Sit, SittingDown(SitD),  Smoking, Waiting (Wait), WalkDog(WalkD), Walk and WalkTogether (WalkT). Best(smallest) values are shown in \textcolor{red}{red} and 2-nd best in \textcolor{blue}{blue}. Methods using Occlusion Aware Training (OAT) are shown with Yes or No, input sequence size is 351 frames.}
    \label{ta:miss16}
    \centering
    \small
    \begin{threeparttable}
    \setlength\tabcolsep{1.2pt} 
    
    \begin{tabular}{|c|c|c|c|c|c|c|c|c|c|c|c|c|c|c|c|c|c|}
    \hline
        \textbf{Method} & \textbf{OAT} &\textbf{Dir.} & \textbf{Disc} & \textbf{Eat} & \textbf{Greet} & \textbf{Phone} & \textbf{Photo} & \textbf{Pose} & \textbf{Purch.} & \textbf{Sit} & \textbf{SitD.} & \textbf{Smoke} & \textbf{Wait} & \textbf{WalkD} & \textbf{Walk} & \textbf{WalkT.} & \textbf{Avg.} \\ \hline
        \multicolumn{18}{|c|}{Protocol 1}\\
        \hline
        
        \textbf{PoseFormer-V2 } &No& 75.6  & 73.87  & 72.27 & 81.00 & 78.31 & 87.43 & 66.22 & 81.28 & 70.80 & 86.94 & 68.85 & 82.09 & 94.16 & 109.33 & 109.17 & 82.5 \\
        \hline

        \textbf{STCFormer } &No& 73.45 &  80.90 & 75.11 & 83.26 & 74.02 & 93.63 & 66.96 & 83.42 & 65.93 & 75.62 & 68.22 & 70.53 & 105.32 & 129.83 & 125.13& 84.76 \\ \hline
       
        \textbf{MHFormer } &No& 71.62 & {73.38} & 71.98 & 79.56 & 71.06 & 85.79 & 64.90 & 80.76 & 63.07 & {75.57} & {64.89} & 80.13 & 93.84 & 121.13 & 123.25 & 81.39 \\ \hline
        
        \textbf{STE} & No&74.82 & 76.49 & 76.33 & 84.54 & 74.45 & 92.31 & 68.63 & 79.87 & 65.76 & 75.83 & 68.26 & 85.67 & 99.00 & 143.84 & 140.32 & 87.08 \\ \hline
       
        \textbf{PoseFormer } &No& 63.02 & 83.09 & 69.60 & 79.29 & {70.10} & 87.13 & 65.59 & 80.23 & \textcolor{blue}{62.78} & 74.85 & 65.83 & {69.60} & 95.01 & 114.21 & 128.95 & {80.60} \\ \hline
        \textbf{Att-3D }&No & 63.08 & 75.37  & 74.96 & 82.99 & 76.22 & 86.33 & 66.20 & 88.09 & 65.36 & 79.44 & 65.65 & 70.31 & 97.08 & 128.89 & 118.22 & 82.6 \\ \hline
        \textbf{VideoPose3D } &No& 64.99 & 75.96 & 74.24 & 81.13 & 76.16 & 91.27 & 68.17 & 84.12 & 66.42 & 78.63 & 66.96 & 71.19 & 99.42 & 133.13 & 125.75 & 83.8 \\ \hline
         \textbf{PoseFormer-V2 } &Yes& 58.51 & 60.00&57.64&64.20&60.22&73.76&58.74&59.59&67.58&74.60&63.04&57.00&68.03&55.46&52.82&62.1
        \\ \hline
          \textbf{Uplift-Upsample } & Yes &{62.72} & 74.35 & 74.63 & 82.52 & 71.69 & {83.36} & 66.48 & 85.07 & 64.46 & 77.11 & 66.37 & 70.35 & 98.41 & 129.88 & 132.90 & 82.69 \\ \hline
        \textbf{P-STMO } & Yes & 70.79 & 73.92 & {73.15} & {79.14} & 70.50 & 86.50 & {64.28} & 85.18 & 64.96 & 74.26 & 65.43 & 80.77 & 95.44 & 125.64 & 127.29 & 82.48 \\ \hline
         \textbf{T3D-CNN } & Yes& 68.68 & 73.98 & 78.15 & 80.1 & 83.25 & 108.33 & 72.86 & {70.94} & 93.21 & 107.77 & 80.83 & 78.58 & {86.7} & {63.25} & {67.67} & 81 \\ \hline 
         \textbf{MHFormer } &Yes& \textcolor{blue}{58.04}&\textcolor{blue}{55.03}&\textcolor{blue}{51.77}&\textcolor{blue}{59.62}&\textcolor{blue}{54.45}&\textcolor{blue}{71.24}&\textcolor{blue}{55.92}&\textcolor{blue}{59.27}&62.95&\textcolor{blue}{72.53}&\textcolor{blue}{54.61}&\textcolor{blue}{60.01}&\textcolor{blue}{63.38}&\textcolor{blue}{47.68}&\textcolor{blue}{48.63}&\textcolor{blue}{58.34} \\ \hline
         \textbf{Ours (DTF)} & Yes&\textcolor{red}{55.81} & \textcolor{red}{54.75} & \textcolor{red}{50.20} & \textcolor{red}{58.40} & \textcolor{red}{53.64} & \textcolor{red}{67.59} & \textcolor{red}{53.19} & \textcolor{red}{55.76} & \textcolor{red}{61.18} & \textcolor{red}{69.40} & \textcolor{red}{53.22} & \textcolor{red}{56.94} & \textcolor{red}{60.86} & \textcolor{red}{45.13} & \textcolor{red}{47.02} & \textcolor{red}{56.21} \\ \hline
        \multicolumn{18}{|c|}{Protocol 2}\\
        \hline
          \textbf{Method}  & \textbf{OAT} & \textbf{Dir.} & \textbf{Disc} & \textbf{Eat} & \textbf{Greet} & \textbf{Phone} & \textbf{Photo} & \textbf{Pose} & \textbf{Purch.} & \textbf{Sit} & \textbf{SitD.} & \textbf{Smoke} & \textbf{Wait} & \textbf{WalkD} & \textbf{Walk} & \textbf{WalkT.} & \textbf{Avg.} \\ \hline
        
        \textbf{PoseFormer-V2 }&No &  55.17 & 57.20  & {57.18} & 64.70 & 56.25 & 63.37 & 51.15 & 59.70 & 54.03 & 66.86 & 53.36 & 57.21 & {72.00} & 86.85 & 85.75 & 62.7 \\
        \hline
         
        \textbf{STCFormer } &No& 61.50 & 66.15  & 60.96 & 68.34 & 57.85 & 68.82 & 53.56 & 65.21 & 53.74 & 61.73 & 55.27 & 55.58 & 83.82 & 104.51 & 99.44& 67.77 \\ \hline
        
        \textbf{MHFormer } &No& 57.43 & 59.15 & 58.44 & 65.65 & 55.67 & 65.53 & 51.55 & 61.40 & 51.55 & 60.61 & 53.14 & 60.55 & 73.27 & 95.12 & 94.71 & 64.25 \\ \hline
        
        \textbf{STE }&No & 58.05 & 62.72 & 60.99 & 69.12 & 58.11 & 70.23 & 55.20 & 63.28 & 53.86 & 61.81 & 55.09 & 62.02 & 78.85 & 109.60 & 105.35 & 68.29 \\ \hline
       
        \textbf{PoseFormer } &No& {51.34} & 62.01 & {56.53} & 64.67 & {54.20} & 64.42 & 51.57 & 59.78 & {50.80} & 59.22 & 52.34 & {53.17} & 73.79 & 90.50 & 94.95 & 62.6 \\ \hline
        \textbf{Att-3D } &No& 52.14 & 59.36 & 59.17 & 67.37 & 56.86 & 64.62 & 51.33 & 64.05 & 52.77 & 62.50 & 52.41 & 54.68 & 73.95 & 98.68 & 89.99 & 64.0 \\ \hline
        \textbf{VideoPose3D }&No & 51.56 & 58.25 & 57.79 & 64.46 & 56.29 & 65.88 & 52.23 & 61.24 & 52.51 & 61.83 & 52.36 & 54.58 & 75.08 & 100.27 & 94.34 & 63.9 \\ \hline
        \textbf{PoseFormer-V2 } &Yes& 46.51 & 47.24&46.55&52.59&46.95&54.04&45.24&44.25&54.32&59.38&49.53&43.65&53.79&43.53&43.33&48.7
        \\ \hline
        \textbf{Uplift-Upsample } &Yes& 50.67 & 57.34 & 58.69 & 65.49 & 54.54 & {62.81} & {51.01} & 62.43 & 52.63 & 61.29 & 52.81 & 54.23 & 74.58 & 96.29 & 99.52 & 63.62 \\ \hline
        \textbf{P-STMO}&Yes & 57.39 & 58.80 & 57.56 & 64.66 & 54.35 & 64.68 & 51.02 & 61.56 & 52.39 & {58.93} & {52.31} & 61.89 & 72.95 & 94.16 & 93.90 & 63.77 \\ \hline
         \textbf{T3D-CNN  
        }&Yes & 53.74 & {57.06} & 62.03 & {63.37} & 62.12 & 76.37 & 57 & {52.39} & 73.14 & 83.53 & 61.55 & 60.37 & {65.48} & {50.94} & {56.26} & {62.4} \\ \hline 
        \textbf{MHFormer } &Yes&  \textcolor{blue}{45.44}&\textcolor{blue}{43.79}&\textcolor{blue}{42.35}&\textcolor{blue}{47.39}&\textcolor{blue}{42.33}&\textcolor{blue}{51.02}&\textcolor{blue}{43.51}&\textcolor{blue}{43.66}&\textcolor{blue}{50.55}&\textcolor{blue}{58.30}&\textcolor{blue}{43.82}&\textcolor{blue}{45.86}&\textcolor{blue}{48.47}&\textcolor{blue}{37.10}&\textcolor{blue}{38.77}&\textcolor{blue}{45.49}\\ \hline
        \textbf{Ours (DTF)} & Yes&\textcolor{red}{44.12} & \textcolor{red}{43.72} & \textcolor{red}{41.76} & \textcolor{red}{47.46} & \textcolor{red}{41.96} & \textcolor{red}{49.96} & \textcolor{red}{41.80} & \textcolor{red}{42.46} & \textcolor{red}{49.19} & \textcolor{red}{55.80} & \textcolor{red}{43.07} & \textcolor{red}{43.78} & \textcolor{red}{47.50} & \textcolor{red}{35.74} & \textcolor{red}{37.62} & \textcolor{red}{44.37} \\ \hline
    \end{tabular}
   
  \begin{tablenotes}\small \footnotesize
\item  \textsuperscript{*}T3D-CNN results in original paper are better because input was ground-truth annotations while here  CPN detections are used.
\end{tablenotes}
\end{threeparttable}  
\end{table*}
\section{Experimental Evaluation} \label{sec:exp}
The proposed DTF algorithm is evaluated  on Human3.6M  \cite{ionescu2013human3} and MPI-INF-3DHP \cite{mehta2017monocular} datasets. Human3.6M is one of the largest datasets with 3.6M frames. It is an indoor environment dataset consisting of 15 different actions performed by 11 subjects. Following previous approaches~\cite{liu2020attention,li2022mhformer,zhang2022mixste}, we have used subjects $\{{1, 5, 6, 7, 8}\}$ for training and subjects $\{{9, 11}\}$  for testing. 
For Human3.6M dataset instead of using 2D ground-truth we have done experiments with 2D CPN detections as input. To demonstrate the occlusion handling capability of the proposed method, we randomly occlude a fixed percentage of these detections. Compared to the 2D ground-truth, the CPN detections have some inherent error which makes the task of 3D estimation more difficult. 
For evaluation, we use MPJPE under Protocol 1 \& 2. Protocol 1 is MPJPE between ground truth and estimated 3D joints. In Protocol 2, we align the estimated 3D poses with the ground-truth poses using scaling, translation, and rotation, which is followed by the calculation of MPJPE. 

MPI-INF-3DHP \cite{mehta2017monocular} is a challenging large scale dataset  consisting of 1.3M frames and offers a diverse range of motions. Following the setting described in \cite{tang20233d,shan2022p,chen2021anatomy,zheng20213d}, we present our evaluation results based on the following metrics:  MPJPE, Percentage of Correct Keypoint (PCK) using a threshold of 150 mm, and Area Under Curve (AUC) across different PCK thresholds.
\begin{table*} 
    \caption{Comparison of existing SOTA methods with random 16 missing joints per frame as input using MPI-INF-3DHP dataset using PCK and AUC measure , with the input sequence size set to 81 frames.}
   \centering
  
    \label{ta:mpi_16}
\begin{tabular}{|c|ccccc||ccccc|}
\hline
& \multicolumn{5}{c||}{PCK $\uparrow$}  & \multicolumn{5}{c|}{\textbf{AUC $\uparrow$}} \\ \hline
\textbf{Activity} & \multicolumn{1}{l|}{\begin{tabular}[c]{@{}l@{}}\textbf{T3D-CNN} \\ \cite{ghafoor2022quantification}\end{tabular}} &\multicolumn{1}{l|}{\begin{tabular}[c]{@{}l@{}}\textbf{P-STMO} \\ \cite{shan2022p}\end{tabular}} & \multicolumn{1}{l|}{\begin{tabular}[c]{@{}l@{}}\textbf{MHFormer} \\ \cite{li2022mhformer}\end{tabular}} & \multicolumn{1}{l|}{\begin{tabular}[c]{@{}l@{}}\textbf{STCFormer} \\ \cite{tang20233d}\end{tabular}} & \multicolumn{1}{l||}{\textbf{DTF}} & \multicolumn{1}{l|}{\begin{tabular}[c]{@{}l@{}}\textbf{T3D-CNN} \\ \cite{ghafoor2022quantification}\end{tabular}} &\multicolumn{1}{l|}{\begin{tabular}[c]{@{}l@{}}\textbf{P-STMO} \\ \cite{shan2022p}\end{tabular}} & \multicolumn{1}{l|}{\begin{tabular}[c]{@{}l@{}}\textbf{MHFormer} \\ \cite{li2022mhformer}\end{tabular}} & \multicolumn{1}{l|}{\begin{tabular}[c]{@{}l@{}}\textbf{STCFormer} \\ \cite{tang20233d}\end{tabular}} & \multicolumn{1}{l|}{\textbf{DTF}} \\ \hline
\textbf{Standing/Walking} & 29.25 & {46.65} & \textcolor{blue}{55.59}  & 35.3  & \textcolor{red}{83.92} & 9.65&21.55  & \textcolor{blue}{28.27} & 15.38  & \textcolor{red}{51.64}  \\ \hline
\textbf{Exercising} & 27.50 & 48.94  & \textcolor{blue}{59.48}  & 43.06   & \textcolor{red}{80.03}  & 9.93&22.21 & \textcolor{blue}{28.94} & 17.78 & \textcolor{red}{43.96}  \\ \hline
\textbf{Sitting} &21.33 & 64.52  & \textcolor{blue}{69.75}   & 59.42 & \textcolor{red}{82.24} &8.41& 38.00 & \textcolor{blue}{39.39}  & 35.63  & \textcolor{red}{48.55} \\ \hline
\textbf{Reaching/Crouching}& 20.14& 50.94  & \textcolor{blue}{65.29}  & 42.07 & \textcolor{red}{86.23} & 8.16&23.6  & \textcolor{blue}{32.82} & 18.18 & \textcolor{red}{48.15} \\ \hline
\textbf{On The Floor}&17.32  & 49.27  & \textcolor{blue}{54.26}  & 46.39   & \textcolor{red}{62.33}  & 6.01&\textcolor{blue}{25.89}  & 25.86  & 25.01  & \textcolor{red}{32.06} \\ \hline
\textbf{Sports}   &23.73 & 27.76   & \textcolor{blue}{37.45}  & 19.99  & \textcolor{red}{65.87}  & 7.35 &9.76  & \textcolor{blue}{14.64}  & 7.06  & \textcolor{red}{32.17}  \\ \hline
\textbf{Miscellaneous}  &25.89& 51.69   & \textcolor{blue}{59.67} & 39.53  & \textcolor{red}{74.78}  &8.65& 24.35  & \textcolor{blue}{29.5}  & 17.99  & \textcolor{red}{41.78}  \\ \hline
\textbf{Average}  & 23.83&50.64  & \textcolor{blue}{59.29}  & 42.28  & \textcolor{red}{77.81}&8.45   & 25.04   & \textcolor{blue}{29.95}   & 20.61 & \textcolor{red}{43.98}    \\ \hline
\end{tabular}
\end{table*}


\subsection{Experimental Settings}
The proposed DTF algorithm is implemented in PyTorch Framework using GeForce RTX 2080 GPU. We divide the occlusion in three different categories including mild (missing joints $\le$ 1/3 joints), intermediate (1/3 $\le$ missing joints$\le$ 2/3 joints) and severe(missing joints $>$ 2/3 joints) occlusion. DTF architecture consists of four layers of MVG followed by two layers of self-refinement module, one layer of information fusion module, and a regression layer at the end to generate a 3D pose. The proposed model is trained end-to-end  using Amsgrad Optimizer. We have used 2D pose detection using Cascaded Pyramid Network (CPN)~\cite{chen2018cascaded} for the Human3.6M dataset and 2D ground-truth for the MPI-INF-3DHP dataset. The size of input 2D pose sequence $P$ is $351 \times 17 \times 3$ for Human3.6M and for MPI-INF-3DHP it is $81 \times 17 \times 3$. For all compared methods, the input 2D pose sequences are temporally interpolated using our proposed occlusion-guidance mechanism \ref{subsec:occ}. Existing SOTA methods are retrained using proposed Occlusion Aware Training (OAT) resulting in significant performance improvement.

\subsection{Quantitative Comparisons}
The proposed DTF algorithm is compared with existing SOTA methods on both datasets. For fair comparisons, the input to all compared methods is the same as obtained by our proposed occlusion guidance mechanism \ref{subsec:occ} except~\cite{ghafoor2022quantification}, due to its independent occlusion handling approach.

\textbf{Comparison on Human3.6M dataset:} Table \ref{ta:miss16} shows action-wise MPJPE comparison with SOTA methods for $16$ out of $17$ joints are missing in each frame for the Human3.6M dataset. This experiment evaluates the proposed algorithm for a very sparse input having $5.88\%$ available joints. The proposed DTF algorithm obtained the average MPJPE of $56.21$mm and $44.37$mm under Protocols 1 \& 2 respectively which is minimum (best) among all compared methods. This experiment demonstrates the occlusion handling capability of the proposed algorithm. We have also compared our proposed method by increasing available joints to $7$ random joints out of $17$ resulting in reduced MPJPE for all compared methods however, the proposed DTF algorithm still achieved the minimum error of $48.35$mm and $38.65$mm under protocols 1 \& 2 respectively as given in Table IX in the supplementary material. We have also evaluated our proposed method with $13$ available joints as shown in Table X in supplemental material. 
Among all existing SOTA methods \cite{pavllo20193d,liu2020attention,zheng20213d,li2022exploiting,li2022mhformer,tang20233d}, we have randomly missed 2D CPN detections and filled input 2D pose sequences with our proposed
occlusion guidance mechanism \ref{subsec:occ}. We have also retrained MHFormer \cite{li2022mhformer}, and PoseFormer-V2 \cite{zhao2023poseformerv2} in an occlusion-aware manner using our proposed occlusion guidance mechanism. Comparison with occlusion-aware SOTA methods \cite{ghafoor2022quantification,shan2022p,einfalt2023uplift,li2022mhformer,zhao2023poseformerv2} also demonstrate improved performance of the proposed DTF algorithm in Table \ref{ta:miss16}.

The performance reported for T3D-CNN \cite{ghafoor2022quantification} in Table \ref{ta:miss16} is lower than that reported in their paper. It is because all experiments in Table \ref{ta:miss16} are performed using CPN detections while the results reported in their paper were using ground-truth. In a new experiment, we performed 3D estimation using the same settings as theirs and we obtained  MPJPE of 31.19mm and 23.40mm under protocol 1 and 2 respectively. In comparison, T3D-CNN has reported much higher values of 39.0mm and 30.4mm under the same protocols.

Among all actions for ``SitD.'' and ``WalkDog'', our proposed DTF algorithm obtained the maximum error due to more complex poses. On the other hand, for  ``Walk'' and ``WalkT'' actions it achieved minimum error which may be attributed to relatively simple poses in these actions. 


\textbf{Comparison on MPI-INF-3DHP dataset:}
Experiments are also performed on the MPI-INF-3DHP dataset \cite{mehta2017monocular} by introducing random omissions of 2D joint positions. We used our occlusion-guidance mechanism \ref{subsec:occ} to fill these missing input 2D pose sequences for all algorithms except \cite{ghafoor2022quantification}. Table \ref{ta:mpi_16} shows the comparison of SOTA methods for random $16$ missing joints out of $17$ per frame in terms of PCK and AUC measures for seven different activities. The proposed DTF algorithm has obtained the best PCK and AUC scores for all activities with severe occlusion. Performance comparisons are also made for random $14$ missing joints out of $17$ per frame. Details can be viewed in the supplementary material.
Due to reduced occlusion, performance of all algorithms has improved, however, the proposed DTF algorithm has still remained the best performer.

\begin{table}[t]
\centering

\caption{Ablation of proposed DTF architecture with varying modules for missing 4 random 2D joints per frame for Protocols 1 \& 2.}
\label{ta:abs}
\begin{tabular}{l|c|l}

\hline
Variant & MPJPE (P1) & MPJPE(P2) \\ \hline
Single View Generator w/o  & &\\Information Fusion (SVG)& 52.95      & 41.24     \\  \hline
Dual View Generator w/o &&\\ Information Fusion (DVG)   & 48.96      & 39.52     \\  \hline
Three View Generator with &&\\Information Fusion (TVG)  & \textcolor{blue}{44.66}      & \textcolor{blue}{35.47}     \\ \hline
Dual View Generator with  &&\\Information Fusion (DTF)    & \textcolor{red}{43.19}      & \textcolor{red}{34.40}     \\ \hline
\end{tabular}
\end{table}
\begin{figure*}[t]

\includegraphics[width=\textwidth,keepaspectratio]{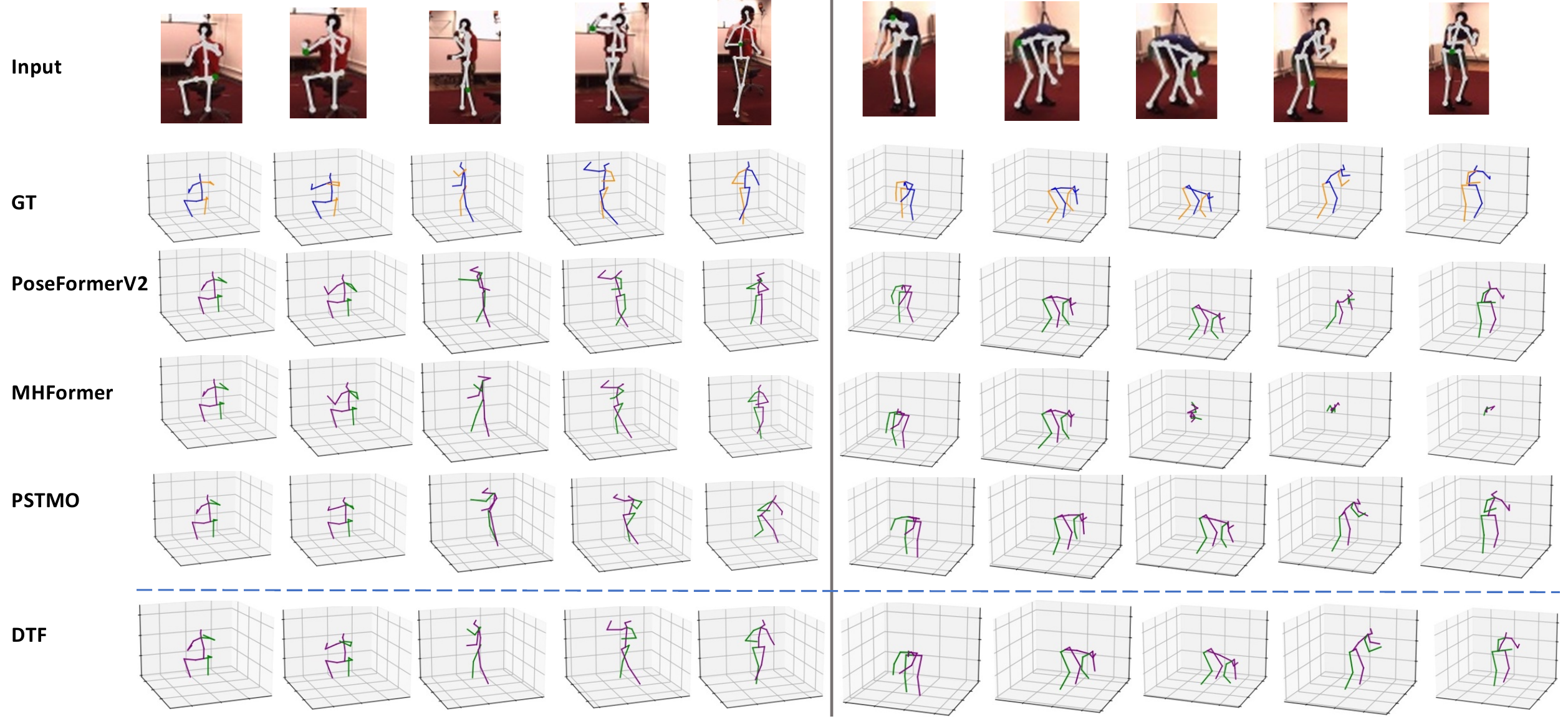}
\caption{Qualitative Comparison of existing SOTA methods with random $16$ missing joints per frame as input using Human3.6M dataset.}
\label{fig:miss16}
\end{figure*}
The comparison of MPJPE under Protocol 1 is shown in Fig. \ref{fig:mpjpe} where the number of random missing joints increases from $4$ to $16$ in a regular interval of $2$. As the number of missing joints per frame is increased, the error for each of the compared algorithms also increases. The proposed algorithm has exhibited a minimum error in the case of severe occlusions i.e. $12, 14,$ \& $16$ missing joints per frame. The performance of the proposed DTF algorithm may be attributed to its novel architecture, self-refinement, and information fusion which attribute to occlusion handling and robustness to the noisy input values generated by the occlusion guidance mechanism.
\subsection{Qualitative Comparisons}
Fig. \ref{fig:miss16} shows comparison of proposed DTF with SOTA on two different actions ``Eating'' and ``WalkDog'' from the Human3.6M dataset for random $16$ missing joints per frame. 3D poses are obtained and compared using various methods including  PoseFormerV2 \cite{zhao2023poseformerv2}, MHFormer \cite{li2022mhformer}, PSTMO \cite{shan2022p} and the DTF algorithm for five frames in each action. The 3D poses recovered by the  DTF algorithm are more closer to the ground-truth 3D poses. Among these methods, MHFormer \cite{li2022mhformer} has shown degenerated results in the case of complex input poses. 
We have also shown visual comparisons on the MPI-INF-3DHP dataset in  supplementary document. The  DTF algorithm showcases excellent performance in both indoor and complex outdoor scenes. 

\subsection{Ablation Study}\label{sec:abl}
We compare different variations of the proposed DTF algorithm including a Single View Generator (SVG) a single transformer architecture without IFM, a Dual View Generator (DVG) with dual transformer architecture without IFM, a Three View Generator (TVG) with three transformer architecture with IFM. Table \ref{ta:abs} shows the comparison of these variants with the proposed DTF  in terms of MPJPE using Protocol 1 \& 2 on the Human3.6M dataset under random $4$ missing joints per frame. The SVG has obtained higher error due to lesser representation power, the DVG obtained significantly reduced error compared to SVG however due to lack of information fusion, its performance remained less than the  DTF algorithm. The TVG version also obtained reduced performance due to a very large number of variables to be learned. These experiments demonstrate the effectiveness of the proposed DTF. More comparisons and ablation experiments such as the effect of the variation of input sequence size as \{81, 243, 351\} on performance are given in the supplementary document. 
\section{Conclusion}
\label{sec:concluions}

In conclusion, this study has pursued the enhancement of 3D Human Pose Estimation (HPE) within challenging scenarios marked by severe occlusion. To this end, we have introduced the Dual Transformer Fusion (DTF) algorithm, an innovative approach grounded in the Transformer framework. This algorithm not only introduces an ingenious occlusion guidance mechanism to adeptly manage sparse 2D input detections but also offers a holistic estimation of multiple views for potential 3D poses. These views are then subjected to refinement via a self-refinement algorithm, collectively elevating the precision of the 3D HPE process. Furthermore, the integration of an information fusion scheme across multiple refined views contributes to an additional improvement in the performance of 3D HPE. Our diligent experimental assessments, conducted across the Human3.6M and MPI-INf-3DHP datasets, substantiate the outstanding capabilities of the proposed DTF algorithm. In navigating the complexities posed by severe occlusion, the DTF algorithm has  demonstrated its prowess, cementing its position at the forefront of 3D HPE achievements. An interesting future work would be to extend the DTF method for multi-person 3D pose estimation  under severe occlusions.

\noindent\textbf{Acknowledgement:}
“This research work was funded by Institutional Fund Projects under grant no. (IFPIP: 1049-135-1443). The authors gratefully acknowledge technical and financial support provided by the Ministry of Education and Deanship of Scientific Research (DSR) at King Abdulaziz University, Jeddah, Saudi Arabia.”

\newpage

\appendices

\section{Example of Occlusion Guidance Mechanism}
In Figure \ref{fig:int}, for demonstration purposes, a search window $f_p=f_f=3$ is shown resulting in a sequence of size $f_p+f_f+1 = 7$ frames. For the frames near the start and the end of the sequence, the past and future windows are adjusted to maintain the overall search window of $f_p + f_f + 1$. For the first frame in the sequence $f_p=0$ and $f_f=6$ while for the last frame $f_p=6$ and $f_f=0$. 
\section{Occlusion Quantization Standardization}
The protocol of randomly eliminating joints has been adopted to standardized the quantization of occlusion in the human pose detection \cite{ghafoor2022quantification,einfalt2023uplift,shan2022p}. This approach facilitates to induce low, medium and sever occlusion in the 2D pose detections and then one may study the response of various algorithms to such intervention. Currently, there is no other standard method proposed for comparing the occlusion handling capability of different 3D pose lifting algorithms. 

Most of the results shown in this work are based on CPN detections which are publicly available and have been used by many SOTA methods. However, the confidence score is not  available with any of the publicly CPN detections datasets. Therefore, it becomes difficult to compare different algorithm by thresholding CPN confidence score to induce occlusion.

To bridge this gap, we consider using stacked-hourglass detections \cite{newell2016stacked} which are publicly available and also contain the confidence score. We analyze the confidence scores as shown in Figures \ref{fig:hist} \& \ref{fig:acc} for Human 3.6M dataset. We observe that \%age of joints having confidence less than 0.6, 0.7, and 0.8 are only $\{4.66\%,16.08\%,37.0\%\}$ which may be considered as mild, intermediate and severe occlusion because in many frames all joints are eliminated while in some frames no joints were missed. For these comparisons with no occlusion using SH-detections is shown in Table \ref{stacked}. The proposed DTF algorithm has obtained quite good results compared to the existing SOTA methods. Figure \ref{fig:cmpd} compares the occlusion handling capability of MHFormer \cite{li2022mhformer} and  PoseFormerV2 \cite{zhao2023poseformerv2} along with our proposed DTF algorithm. This comparison also shows the improved occlusion handling capability of proposed DTF algorithm.

\setcounter{figure}{3}
\begin{figure*}
\includegraphics[width=\textwidth]{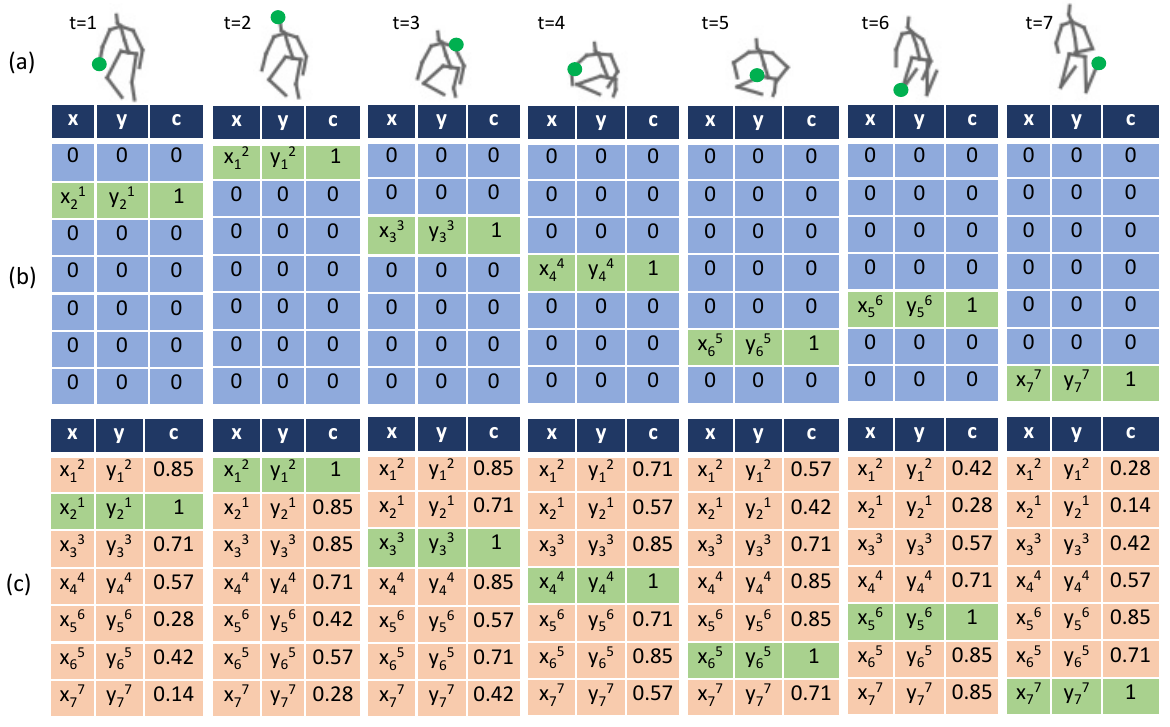}
\caption{(a) A sequence of seven detected 2D skeletons where the green marker shows the available joint in each frame. (b) 2D joint positions with confidence $(x,y,c)$, where (0,0,0) show missing joints. (c) Missing joint positions are interpolated and a confidence score is assigned based on the temporal gap.}
\label{fig:int}
\end{figure*}

\section{Quantitative Comparison on MPI-INF-3DHP Dataset}

MPI-INF-3DHP dataset has more complex poses with occlusion. Table \ref{ta:mpi_14} shows performance comparisons for random $14$ missing joints out of $17$ per frame under PCK and AUC as performance measures. The results showcase the effectiveness of the proposed DTF algorithm in accurately estimating 3D poses even under severe occlusion conditions.

\setcounter{table}{3}
\begin{table*}
    \caption{Comparison of existing SOTA methods with random 14 missing joints per frame as input from MPI-INF-3DHP dataset using PCK and AUC measure\textcolor{blue}{, with the input sequence size set to 81 frames.}}
    \setlength\tabcolsep{0pt}
     \small
     
    \label{ta:mpi_14}
\begin{tabular}{|c|ccccc||ccccc|}
\hline
& \multicolumn{5}{c||}{PCK $\uparrow$}  & \multicolumn{5}{c|}{\textbf{AUC $\uparrow$}} \\ \hline
\textbf{Activity} & \multicolumn{1}{l|}{\begin{tabular}[c]{@{}l@{}}\textbf{T3D-CNN} \cite{ghafoor2022quantification} \\ \end{tabular}} &\multicolumn{1}{l|}{\begin{tabular}[c]{@{}l@{}}\textbf{P-STMO \cite{shan2022p}} \\ \end{tabular}} & \multicolumn{1}{l|}{\begin{tabular}[c]{@{}l@{}}\textbf{MHFormer \cite{li2022mhformer}} \\ \end{tabular}} & \multicolumn{1}{l|}{\begin{tabular}[c]{@{}l@{}}\textbf{STCF \cite{tang20233d}} \\ \end{tabular}} & \multicolumn{1}{l||}{\textbf{DTF}} & \multicolumn{1}{l|}{\begin{tabular}[c]{@{}l@{}}\textbf{T3D-CNN \cite{ghafoor2022quantification}} \\ \end{tabular}} &\multicolumn{1}{l|}{\begin{tabular}[c]{@{}l@{}}\textbf{P-STMO \cite{shan2022p}} \\ \end{tabular}} & \multicolumn{1}{l|}{\begin{tabular}[c]{@{}l@{}}\textbf{MHFormer \cite{li2022mhformer}} \\ \end{tabular}} & \multicolumn{1}{l|}{\begin{tabular}[c]{@{}l@{}}\textbf{STCF \cite{shan2022p}} \\ \end{tabular}} & \multicolumn{1}{l|}{\textbf{DTF}} \\ \hline

\textbf{Standing/Walking} &14.27  & 80.36  & \textcolor{blue}{86.72} & 66.23  & \textcolor{red}{91.66}  &4.91& 47.86 & \textcolor{blue}{53.57}  & 34.31  & \textcolor{red}{57.50}  \\ \hline
\textbf{Exercising} & 12.78& 82.36  & \textcolor{blue}{87.16}  & 72.72   & \textcolor{red}{89.24}  &3.88 &50.23 & \textcolor{blue}{52.94} & 39.92 & \textcolor{red}{55.32}   \\ \hline
\textbf{Sitting}&13.84&  \textcolor{blue}{88.83}  & 89.56 & 81.40 & \textcolor{red}{90.76} &3.83 &\textcolor{red}{58.19}  & 55.40  & 53.13  & \textcolor{blue}{56.79}\\ \hline
\textbf{Reaching/Crouching}&15.37 & 81.57 & \textcolor{blue}{89.95}   & 70.52   & \textcolor{red}{92.20}  & 6.68& 48.80 & \textcolor{blue}{54.62}  & 37.93   & \textcolor{red}{56.06} \\ \hline
\textbf{On The Floor}  & 13.75&\textcolor{red}{80.55}   & 57.29   & 72.20 & \textcolor{blue}{73.23}   & 4.20& \textcolor{red}{47.98}   & 30.55  & \textcolor{blue}{42.79}  & 40.07  \\ \hline
\textbf{Sports} &11.89 & 55.23   & \textcolor{blue}{72.26} & 39.74 & \textcolor{red}{79.61} &4.75& 25.70  & \textcolor{blue}{37.43} & 16.38  & \textcolor{red}{43.42} \\ \hline
\textbf{Miscellaneous} &13.09& 76.5 & \textcolor{blue}{80.79}  & 62.07   & \textcolor{red}{85.67} &3.64 &45.03  & \textcolor{blue}{47.22}  & 32.85  & \textcolor{red}{50.38}  \\ \hline
\textbf{Average} &13.68 & 79.34 & \textcolor{blue}{82.26} & 67.69  & \textcolor{red}{87.12}   &4.49& 47.64   & \textcolor{blue}{48.90}  & 37.85 & \textcolor{red}{52.40}   \\ \hline
\end{tabular}
\end{table*}

\section{Quantitaive Comparison on Human3.6M Dataset}
We have provided additional results with random missing joint sequences as input. For fair comparisons, the input to all the compared methods is the same as obtained by our proposed occlusion guidance mechanism, except T3D-CNN, as it has proposed its own occlusion handling method. Table \ref{ta:miss14} provides a performance comparison of the $10$ random missing joints in each video frame. The proposed DTF algorithm has achieved the error $45.56 mm$ and $36.20 mm$ under protocols 1 \& 2 respectively, significantly outperforming all the compared methods. Table \ref{ta:miss4} shows the comparison of SOTA methods for only $4$ out of $17$ random missing joints in each frame from Human3.6M dataset. In Table \ref{ta:miss4}, all the compared methods have shown reduced errors due to a very small number of missing joints. The proposed DTF algorithm has still obtained the best performance in terms of minimum MPJPE under protocols 1 \& 2.

\textcolor{blue}{Most of the results shown in this work are based on CPN detections which are publicly available and have been used by many SOTA methods. We have introduced occlusion by randomly missing joints in each frame and filled joint positions using proposed occlusion guidance mechanism followed by Dual Transformer Fusion Architecture that have shown improved results as compared to existing SOTA methods.}

\begin{table}[h]
\centering
\caption{ Quantitative Evaluation of the Proposed DTF Architecture on SH-Detections with No Occlusion, Compared to Existing State-of-the-Art (SOTA) Methods.}
\label{stacked}
\begin{tabular}{c|c|c|c}
\hline
Method & $T$ & MPJPE (P1) & MPJPE (P2) \\ \hline
MHFormer \cite{li2022mhformer} & 351 &50.47&40.49\\
PoseFormer-V2 \cite{zhao2023poseformerv2} &243 & 59.32 & 48.64 \\
\hline
DTF & 351 &49.27&40.31\\ \hline
\end{tabular}

\end{table}

\begin{figure}

\includegraphics[width=0.5\textwidth,keepaspectratio]{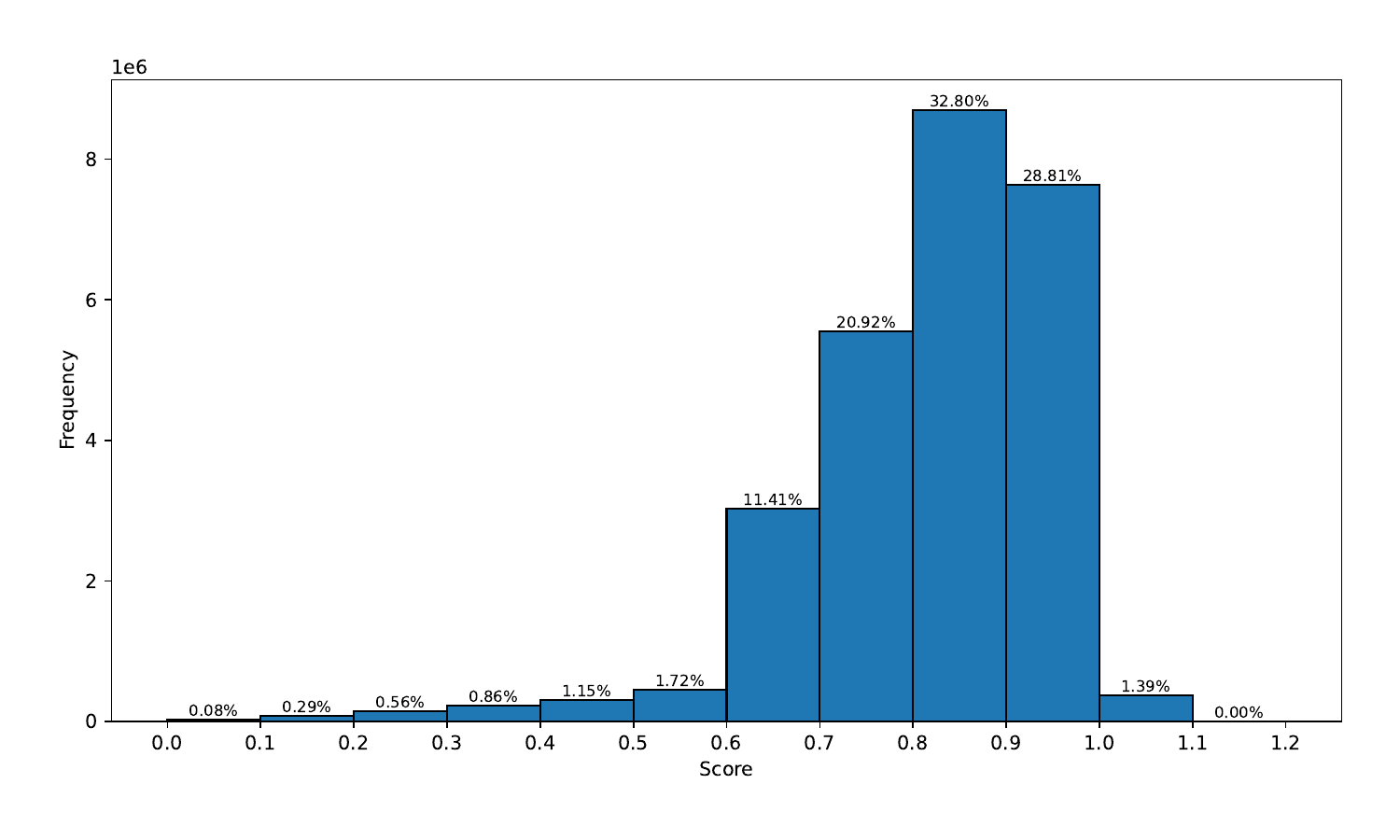}
\caption{Histogram of Confidence Scores for SH-Detections in the Human3.6M Dataset.}
\label{fig:hist}
\end{figure}
\begin{figure}

\includegraphics[width=0.5\textwidth,keepaspectratio]{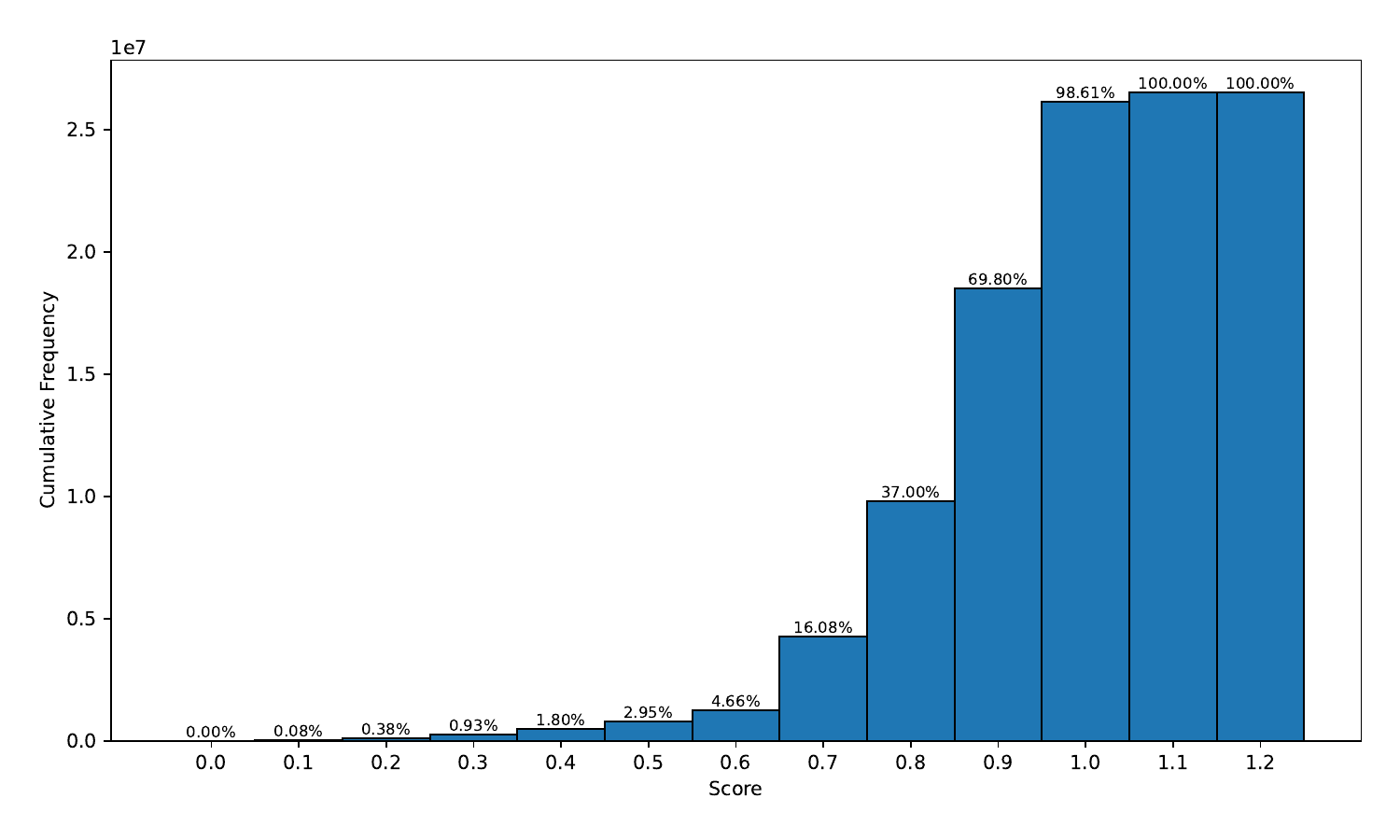}
\caption{ Accumulation Histogram of Confidence Scores for SH-Detections in the Human3.6M Dataset}
\label{fig:acc}
\end{figure}
\begin{figure}
\includegraphics[width=0.5\textwidth,keepaspectratio]{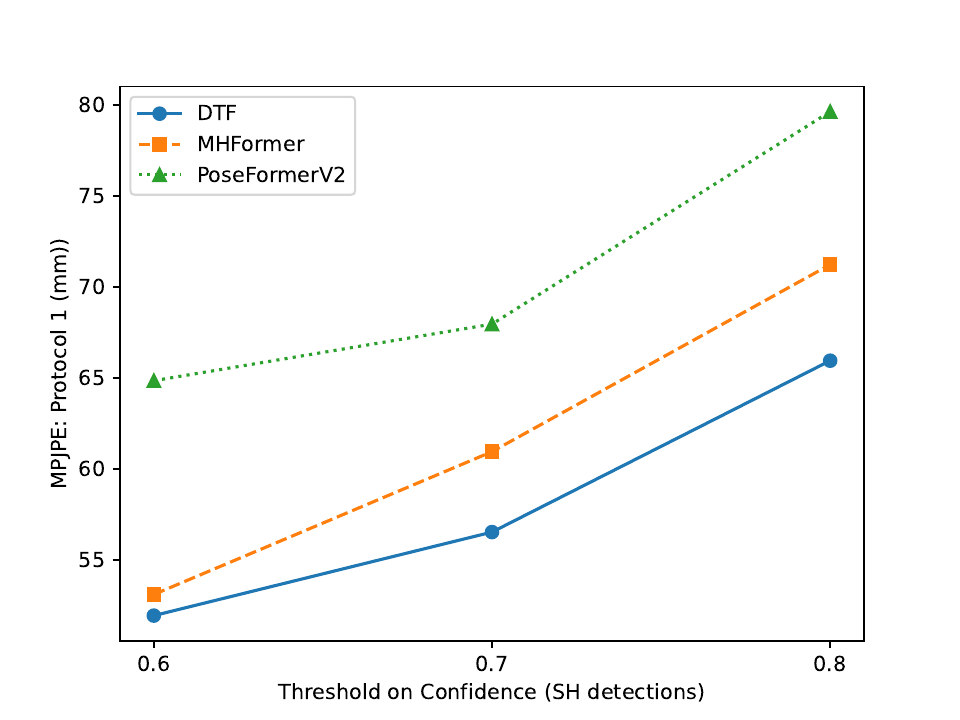}
\caption{Comparison of Proposed DTF algorithm with MHFormer and PoseFormerV2 in terms of MPJPE (mm) under Protocol 1 using 2D SH-Detections from Human3.6M dataset. 2D joints with confidence less than 0.6, 0.7, 0.8 are considered as occluded.}
\label{fig:cmpd}
\end{figure}
\section{Occlusion Categories}
We divide the occlusion in three different categories based on the number of occluded joints including mild, intermediate and severe. Mild occlusion refers to missing less than one-third joints (in Human3.6M dataset 1-5 joints may be missing). Intermediate occlusion refers to missing more than one third but less than two third joints (in Human3.6M dataset 6-11 joints may be missing). Severe occlusion refers to missing more than two third joints (in Human3.6M dataset 12-16 joints may be missing out of total 17 joints). For each occlusion category, we train a single network and we report results for including  4, 10, and 16 missing joints.

\section{Zero Occlusion Comparisons}
We have also reported performance of proposed DTF algorithm without any occlusion in Table \ref{tab:noOcclusion}. Many training parameters i.e. batch size effects the final performance numbers. Higher batch sizes such as 128 and 256 as used by many SOTA methods require more GPU memory and also more computational power. Using a smaller batch size incurs less memory overhead while performance may also reduce. 

In Table \ref{tab:noOcclusion}, the proposed DTF algorithm is compared with MHFormer \cite{li2022mhformer}, PoseFormerV2 \cite{zhao2023poseformerv2}, STE \cite{li2022exploiting}, T3D-CNN \cite{ghafoor2022quantification}, ATT-3D \cite{liu2020attention}, and VideoPose3D \cite{pavllo20193d} using varying batch sizes and sequence sizes. On the average, the proposed DTF algorithm has obtained best performance with a batch size of 32 and sequence size of 351. Therefore, the proposed DTF algorithm not only handles the severe occlusion but also achieve excellent performance in zero occlusion cases.

We have studied the variations of performance of MHFormer \cite{li2022mhformer}, PoseFormerV2 \cite{zhao2023poseformerv2}, Uplift-UpSample \cite{einfalt2023uplift}, and the proposed DTF algorithm for zero occlusion in Table \ref{tab:noOcclusion} and with four random missing joints in Table \ref{ta:frame}. In all of these experiments, we observe performance improvement (MPJPE Error Reduction) with the increase in sequence size across all algorithms. It is because as sequence size increase more contextual information is available. In these experiments, sequence sizes are varied as 81, 243, and 351. In Table \ref{ta:frame} the proposed DTF algorithm has obtained a performance of 43.19/34.40 MPJPE for a sequence size of 351 which is the best performance compared to all methods. In Table \ref{tab:noOcclusion}, the proposed DTF has obtained 43.02/30.39 with no occlusion scenario which is again the best performance across all compared methods.

\section{Ablation on Input Sequence Size Variation}

We have trained our networks by varying the number of frames in a sequence starting from
81, 243 and 351 sequence sizes, following some of the SOTA methods. The results are shown in Table \ref{ta:frame}. The value of 351 frames has been recommended by MHFormer \cite{li2022mhformer} and Uplift-UpSample \cite{einfalt2023uplift}. For fair comparison, we kept the same sequence size as used by these approaches. Selecting different sequence size would change the input size requiring retraining of the original network. As predicted an increase in sequence size results in improved performance across all compared methods.
\begin{table}[t]
\caption{Comparison of Proposed DTF architecture with SOTA methods under random 4 occluded joints by varying sequence size on Human3.6M dataset. MPJPE is reported for Protocol 1/ Protocol 2}
\label{ta:frame}
\begin{tabular}{c|ccc}
\hline
\textbf{Seq Size}& \textbf{81} & \textbf{243} & \textbf{351} \\ \hline
\textbf{Methods:} & & &\\ 
STCFormer &    46.60/36.68   &   -      &      -       \\ 
MHFormer & 47.47/37.32   &46.36/36.61 & 46.60/36.63\\
Uplift-Upsam. & 49.56/38.63  &- &  45.60/36.14\\
P-STMO&-&47.52/36.64&-\\
STE&-&-&56.56/46.49\\
PoseFormer-V2 &47.5/37.0 &46.4/36.4&-\\
PoseFormer &46.2/35.8 &-&- \\
T3D-CNN &-&52.3/40.4&-\\ \hline
Ours (DTF)  & 46.39 /36.42      & 44.0/35.03         &      43.19/34.40  \\ \hline

\end{tabular}
\end{table}

\section{Occlusion-aware Retraining of SOTA Methods}
In order to validate the importance of the proposed occlusion guidance mechanism, occlusion-aware retraining of three SOTA methods including Strided Tranformer (STE) \cite{li2022exploiting}, PoseFormerV2 \cite{zhao2023poseformerv2},
and MHFormer \cite{li2022mhformer} as shown in Table \ref{tab:supp}. In these experiments, we observe significant performance improvement by introducing our proposed occlusion guidance mechanism across all methods. These results demonstrate the applicability of the proposed occlusion guidance approach to any existing SOTA methods. However, the proposed DTF algorithm remained the most accurate. The nearest competitor is MHFormer \cite{li2022mhformer} which obtained 58.34 for protocol 1 and 45.49 for protocol 2 for  the case of random missing 16 joints out of 17. The performance of MHFormer \cite{li2022mhformer} has remained less than DTF algorithm because of large number of parameters with some extent it overfits on occlusion-aware data. 

\section{Qualitative Comparison on MPI-INF-3DHP Dataset}
\begin{figure*}
\includegraphics[width=\textwidth,keepaspectratio]{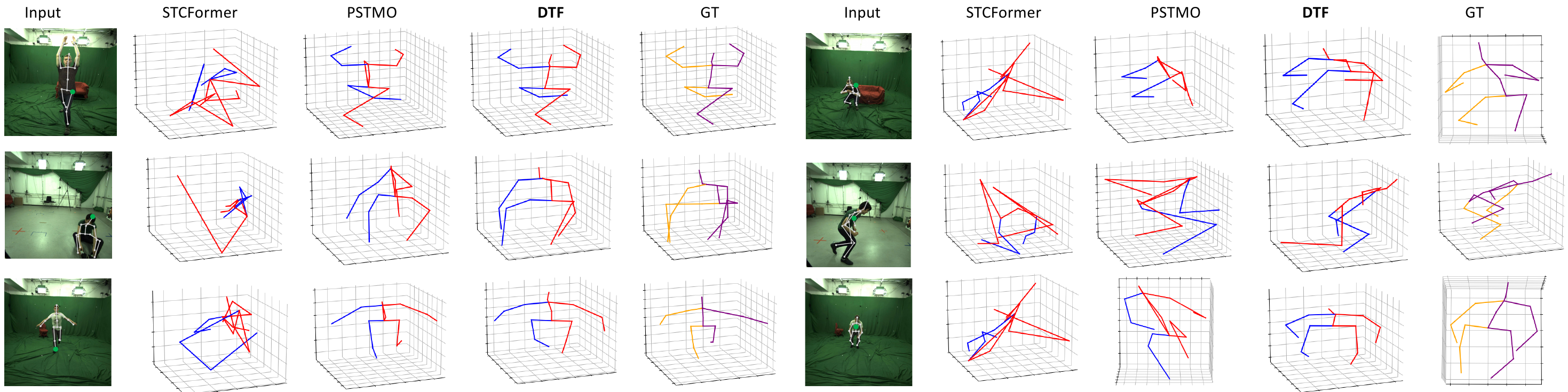}
\caption{Qualitative Comparison of existing SOTA methods with random 16 missing joints per frame as input using MPI-INF-3DHP dataset.}
\label{fig:mpi_16}
\end{figure*}

We have shown visualization comparison on the MPI-INF-3DHP dataset with SOTA on $16$ missing joints per frame in Figure \ref{fig:mpi_16}.

\section{Qualitative Comparison on Human3.6M Dataset}
\begin{figure*}[t]
\includegraphics[width=\textwidth,keepaspectratio]{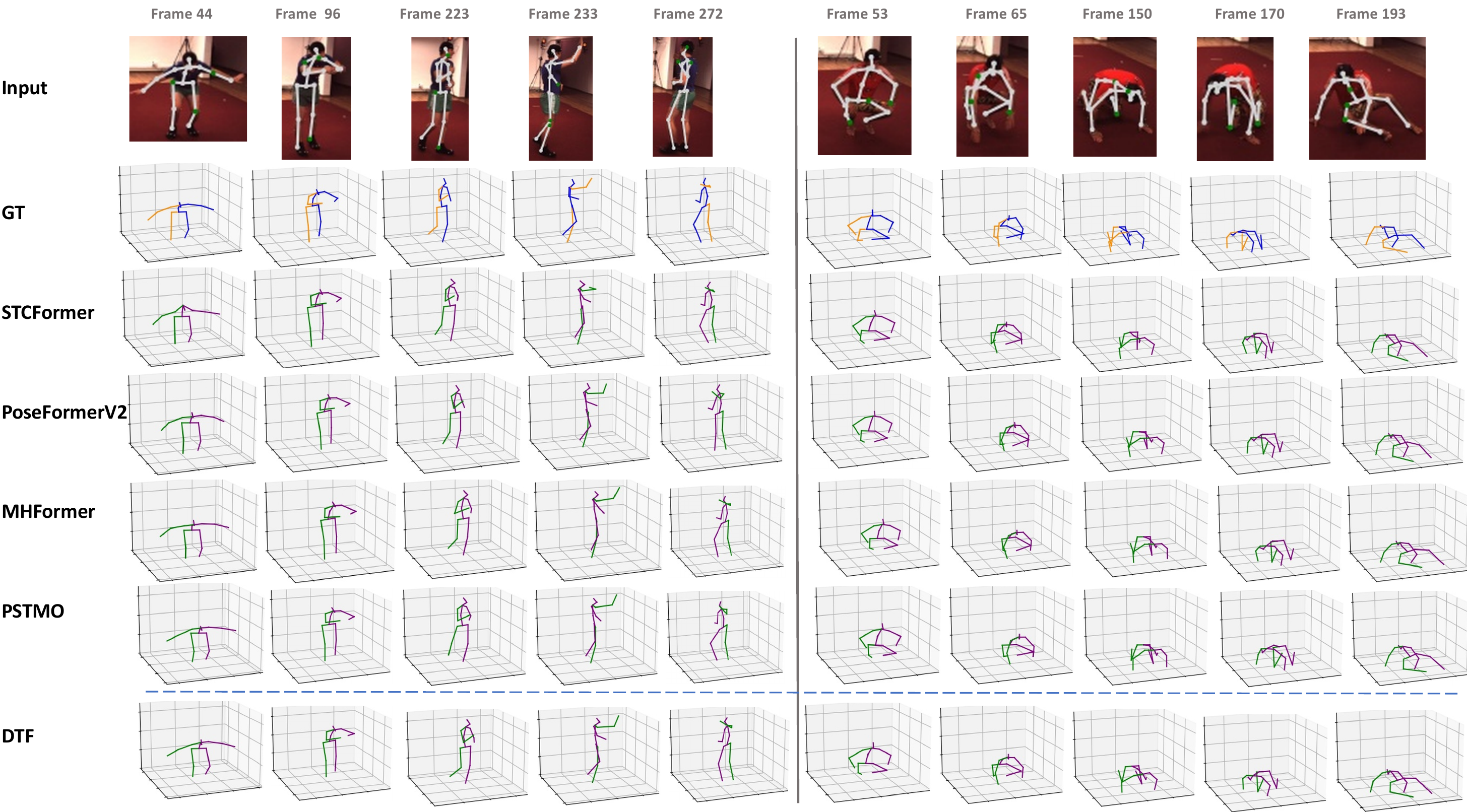}
\caption{Qualitative Comparison of existing SOTA methods with random $14$ missing joints per frame as input using Human3.6M dataset.}
\label{fig:miss14}
\end{figure*}

We have extended the evaluation to include a visualization comparison on Human 3.6M dataset, by inducing the absence of random 14 joints out of the total 17. Figure \ref{fig:miss14} shows a comparison of the proposed DTF with existing state-of-the-art methods by missing $14$ joints out of $17$ from the Human3.6M dataset on actions: ``Sitting Down" and ``Greeting". The results showcase the effectiveness of the proposed
DTF algorithm in accurately estimating 3D poses even under severe occlusion conditions.

\begin{table}[t]
\caption{Comparison of the proposed DTF algorithm with existing methods including STE \cite{li2022exploiting}, MHFormer \cite{li2022mhformer}, and PoseFormer-V2 \cite{zhao2023poseformerv2} with random 16 (out 17) missing joints in terms of MPJPE under Protocol 1 / Protocol 2.  No Occlusion Guidance (NOG) means missing joints are represented with zeros. Occlusion Guidance at Validation (OGV) for missing joints estimation without retraining.  Occlusion Aware Training (OAT) means occlusion guidance is used for retraining as well as during  validation.}
\label{tab:supp}
\begin{tabular}{c|ccc}
\hline
Method  & NOG & OGV & OAT\\\hline
PoseFormer-V2   &443.6/209.6 &82.5/62.7 &62.1/48.7         \\
STE     & 518.88/232.30 &87.08/68.29 &59.05/46.49   \\
MHFormer   & 790.52/413.07 &81.39/64.25 & 58.34/45.49             \\
Ours (DTF)    & -&- & 56.21/44.37  \\ \hline
\end{tabular}
\end{table}

\begin{table}[t]
\caption{Comparison of Proposed DTF architecture with zero occlusion using CPN 2D detections on Human 3.6M dataset}
\label{tab:noOcclusion}
\centering
\begin{tabular}{c|ccc}
\hline
\textbf{Method} & \textbf{Batch Size} & \textbf{Sequence Size} & MPJPE(mm) \\ \hline
MHFormer    & 256 & 81 & 44.5/-   \\ 
MHFormer  & 256& 243 & 43.2/-   \\ 
MHFormer   & 256 & 351  & 43.0/30.5   \\ 
PoseFormer-V2& 1024& 81& 46.0/36.1   \\ 
PoseFormer-V2  & 1024  & 243  & 45.2/35.6   \\ 
STE & 256  & 351 & 43.7/35.2   \\ \
T3D-CNN & 1024  & 243 & 46.8/36.5   \\ \
Att-3D  & 1024   & 243 & 45.1/35.6   \\ 
VideoPose3D  & 1024 & 243  & 46.8/36.5   \\ \hline
Ours (DTF)  & 32 & 351  & 43.02/30.39   \\ \hline
\end{tabular}
\end{table}
\section{Limitations of Proposed Occlusion Guidance Mechanism}
\subsection{Occlusion Guidance Mechanism for Low frame Rates:}
The commonly used datasets in human pose estimation such as Human3.6M and MPI-INF-3DHP datasets have a frame rate of 50fps. To remain consistent with existing SOTA methods, we have used the same frame rate.  The results will not remain comparable across varying frame rates. The proposed occlusion guidance scheme will remain applicable for normal videos having frame rate of 30fps. For lower frame rate i.e. 15fps, the motion between same joints in adjacent frames will not remain smooth.  Therefore existing sequence based pose up-lift methods will face challenges in 3D estimation.

\subsection{Occlusion Guidance Mechanism for Multi-person Pose Estimation using Single View:}  
Most of the existing SOTA pose uplift methods are using single person pose estimation using single 2D view. All the compared methods in this work are single person single view. The proposed occlusion guidance mechanism can be easily modified for multi-person non-crowded scenes. However, if the scene becomes crowded then all pose up-lift methods will face significant difficulty in disambiguating joints of different individuals. In such cases, the proposed occlusion guidance algorithm will also face difficulty.

\subsection{3D Structure Recovery Specific to Human Body:}

We have introduced an end-to-end trainable network DTF to improve the estimation of 3D human body joint locations, when the 2D joint positions are available. Our study  limits its scope to focus on only the human body structure. This means that while our approach effectively addresses human pose estimation challenges, it may not cover all potential issues associated with generic 3D estimation from 2D images.

\begin{table*}[t]

   \caption{Comparison in terms of MPJPE of the proposed DTF algorithm with existing SOTA methods for $10$ random missing joints per frame in Human3.6M dataset using 2D CPN detections as input. Action-wise performance is reported for each algorithm. Methods using Occlusion Aware Training (OAT) are shown with Yes or No. With Input Sequence Size set to 351 Frames.}
   \label{ta:miss14}
    \raggedleft
    \small
    
    \setlength\tabcolsep{1.2pt} 
    \begin{tabular}{|c|c|c|c|c|c|c|c|c|c|c|c|c|c|c|c|c|c|}
    
    \hline
        \textbf{Method} & \textbf{OAT}&\textbf{Dir.} & \textbf{Disc} & \textbf{Eat} & \textbf{Greet} & \textbf{Phone} & \textbf{Photo} & \textbf{Pose} & \textbf{Purch.} & \textbf{Sit} & \textbf{SitD.} & \textbf{Smoke} & \textbf{Wait} & \textbf{WalkD} & \textbf{Walk} & \textbf{WalkT.} & \textbf{Avg.} \\ \hline
        \multicolumn{18}{|c|}{Protocol 1}\\
        \hline
         \textbf{PoseFormerV2 } &No& 50.60 & 56.79  & 49.62 & 50.33 & 59.53 & 61.40 & 46.63 & 48.08 & 63.84 & 76.83 &\textcolor{red}{40.34} & 55.31 & 57.98 & 41.05 & 51.96& 54.9 \\ \hline
        
        \textbf{STCFormer } & No&\textcolor{red}{39.71}&69.39&\textcolor{red}{ 39.47}&{ 42.60}&\textcolor{red}{ 47.48}&59.04&\textcolor{blue}{ 41.18}&\textcolor{red}{ 40.28}&\textcolor{blue}{ 54.88}&63.94&51.70&54.81&53.28&42.87&44.83&49.70 \\ \hline

        \textbf{MHFormer } & No&45.43&48.94&{ 40.45}&\textcolor{red}{41.44}&55.63&53.21&\textcolor{red}{ 40.85}&41.75&66.45&\textcolor{blue}{ 60.45}&52.84&51.20&\textcolor{blue}{ 46.27}&50.05&40.98&{ 49.06} \\ \hline

        \textbf{STE} & No&48.28&52.40&41.56&43.78&64.21&55.77&42.86&41.83&72.23&60.96&54.81&55.42&48.81&65.81&49.06&53.19\\ \hline
        
        \textbf{PoseFormer } &No& 48.78&68.68&\textcolor{blue}{40.17}&{ 43.00}&55.27&58.13&42.44&42.78&\textcolor{red}{ 54.33}&70.70&47.58&\textcolor{blue}{ 43.71}&61.85&34.89&\textcolor{blue}{37.53}&50.0\\ \hline

        \textbf{Att-3D } &No& 49.90&68.00&44.15 &47.71&51.11&60.95&42.99&42.66&65.63&66.98&47.58&\textcolor{red}{43.66}&48.30&38.23&43.71&50.8\\ \hline

        \textbf{VideoPose3D } & No&51.97&55.15&44.76&49.09&60.34&59.18&46.21&45.12&63.26&71.95&49.70&65.9&52.21&49.92&39.04&53.55\\ \hline

         \textbf{Uplift-Upsample } & Yes& 48.27&64.41 &42.92&44.90&58.77&54.64&44.24&43.71&59.81&73.14&{46.49}&53.19&54.01&\textcolor{red}{ 32.45}&37.62&50.57\\ \hline
        \textbf{PoseFormerV2 } & Yes&44.84&48.64&44.41&46.73&50.15&55.24&45.84&44.84&56.92&64.63&49.22&47.32&49.67&35.62&35.36&48.0\\ \hline
         \textbf{MHFormer } & Yes&43.27&\textcolor{blue}{46.42}&42.85&44.30&50.14&56.69&46.01&43.47&59.37&64.61&48.35&46.23&47.23&35.54&35.26&\textcolor{blue}{47.32} \\ \hline
        \textbf{P-STMO } & Yes&46.31&50.00&41.06&\textcolor{blue}{41.88}&59.08&\textcolor{red}{ 52.33}&41.26&\textcolor{blue}{ 40.56}&69.69&\textcolor{red}{ 59.52}&55.01&54.31&\textcolor{red}{ 45.94}&54.62&42.94&50.29 \\ \hline
       
        \textbf{T3D-CNN } &Yes& 53.75&55.49 &57.11&56.81&60.62&75.66&54.82&55.77&72.14&79.66&59.22&57.24&61.5&42.09&41.81&58.9\\ \hline

        \textbf{Ours (DTF)} & Yes&\textcolor{blue}{41.80} & \textcolor{red}{45.58} & {40.76} & {43.04} & \textcolor{blue}{47.73} & \textcolor{blue}{53.16} & {42.69} & {43.53} & {56.52} & {63.04} & \textcolor{blue}{46.16} & \textcolor{blue}{44.71} & {46.46} & \textcolor{blue}{34.71} & \textcolor{red}{33.51} & \textcolor{red}{45.56} \\ \hline
        \multicolumn{18}{|c|}{Protocol 2}\\
        \hline
          \textbf{Method} & \textbf{OAT}&\textbf{Dir.} & \textbf{Disc} & \textbf{Eat} & \textbf{Greet} & \textbf{Phone} & \textbf{Photo} & \textbf{Pose} & \textbf{Purch.} & \textbf{Sit} & \textbf{SitD.} & \textbf{Smoke} & \textbf{Wait} & \textbf{WalkD} & \textbf{Walk} & \textbf{WalkT.} & \textbf{Avg.} \\ \hline
       
        \textbf{PoseFormerV2 } & No&36.78 & 41.34  & 37.52 & 39.65 & 41.79 & 43.74 & 34.79 & 35.63 & 47.70 & 58.89 & 40.34 & 39.03 & 43.98 & 30.53 & 37.96 & 40.6 \\ \hline
       \textbf{STCFormer } & No& \textcolor{red}{ 31.51}&56.55&\textcolor{red}{ 32.35}&34.79&\textcolor{red}{36.71}&45.50&\textcolor{blue}{32.11}&\textcolor{red}{31.41}&\textcolor{blue}{43.79}&51.03&41.46&43.01&41.40&35.19&37.30&39.61\\ \hline

        \textbf{MHFormer } & No&35.70&38.33&33.09&\textcolor{red}{34.34} &41.54&\textcolor{red}{40.28}&\textcolor{red}{ 32.10}&{32.64}&51.57&\textcolor{blue}{49.02}&41.68&38.62&\textcolor{red}{36.09}&36.59&31.62&\textcolor{blue}{ 38.25}\\ \hline
       
         \textbf{STE} & No& 40.14&42.21&33.58&36.64&48.41&44.39&33.95&33.12&55.75&{50.33}&44.22&43.76&39.04&48.88&38.49&42.19\\ \hline
          \textbf{PoseFormer } &No& 36.76&47.14&\textcolor{blue}{32.89}&35.0&40.16&42.88&32.53&\textcolor{blue}{32.54}&\textcolor{red}{43.43}&53.74&\textcolor{blue}{37.94}&\textcolor{red}{32.78}&43.77&\textcolor{blue}{26.38}&{ 28.87}&37.8\\ \hline
          \textbf{Att-3D } &No&38.45&51.38&35.13 &38.92&38.93&46.36&33.40 &33.04&51.25&53.04&38.12&\textcolor{blue}{34.05}&37.92&29.53&34.17&39.6\\ \hline
        \textbf{VideoPose3D } & No&38.92&41.99&35.46&39.72&44.60&45.01&35.43&34.61&49.45&56.59&39.04&47.69&40.04&37.93&30.77&41.20\\ \hline
          \textbf{Uplift-Upsample } & Yes& 38.17&50.05&35.36&37.07&44.29&42.89&34.39&33.79&47.93&56.98&\textcolor{red}{37.74}&41.01&41.88&\textcolor{red}{26.33}&29.70&39.84\\ \hline
          
          \textbf{PoseFormerV2 } & Yes& 34.76&38.07&36.05&37.71&38.74&42.48&34.99&34.20&46.56&52.49&39.68&39.62&38.57&27.74&28.41&37.8\\ \hline
            \textbf{MHFormer } & Yes& 34.47&37.11&35.83&36.18&39.21&43.36&34.80&33.90&47.85&52.78&39.41&35.68&37.26&27.95&\textcolor{blue}{28.02}&\textcolor{blue}{37.59}\\ \hline
         \textbf{P-STMO } & Yes& 35.01&38.61&33.30&3\textcolor{blue}{ 4.59}&41.26&40.65&32.84&32.81&52.21&\textcolor{red}{48.37}&41.70&38.69&\textcolor{blue}{36.22}&36.70&30.76&38.20\\ \hline
          \textbf{T3D-CNN } &Yes& 40.76&43.42&44.83& 45.14&45.51&54.18&42.49&40.89&55.86&63.16&45.47&42.35&46.38&31.18&34.06&45.1\\ \hline
         \textbf{Ours (DTF)} & Yes&\textcolor{blue}{32.84} & \textcolor{blue}{36.40} & {33.33} & {35.26} & \textcolor{blue}{37.26} & \textcolor{blue}{40.44} & {33.07} & {33.22} & {45.64} & {51.28} & {37.95} & {34.71} & {36.44} & {27.83} & \textcolor{red}{27.32} & \textcolor{red}{36.20} \\ \hline

    \end{tabular}
\end{table*}


\begin{table*}[t]
    \caption{Comparison in terms of MPJPE of the proposed DTF algorithm with existing SOTA methods for $4$ random missing joints per frame in Human3.6 M dataset using 2D CPN detections as input. Action-wise performance is reported for each algorithm. Methods using Occlusion Aware Training (OAT) are shown with Yes or No. With Input Sequence Size set to 351 Frames.}
    \label{ta:miss4}
    \raggedleft
    \small
    \setlength\tabcolsep{1.2pt} 
    \begin{tabular}{|c|c|c|c|c|c|c|c|c|c|c|c|c|c|c|c|c|c|}
    
    \hline
        \textbf{Method} & \textbf{OAT}&\textbf{Dir.} & \textbf{Disc} & \textbf{Eat} & \textbf{Greet} & \textbf{Phone} & \textbf{Photo} & \textbf{Pose} & \textbf{Purch.} & \textbf{Sit} & \textbf{SitD.} & \textbf{Smoke} & \textbf{Wait} & \textbf{WalkD} & \textbf{Walk} & \textbf{WalkT.} & \textbf{Avg.} \\ \hline
        \multicolumn{18}{|c|}{Protocol 1}\\
        \hline
       \textbf{PoseFormerV2 } & No&44.01 &{ 47.22} & 41.48 & 44.09 & 48.39 & 53.76&42.68 & 42.66 & 57.76 & 64.65 & 45.75 & 50.37&46.89&32.40&{ 34.22}&46.4\\ \hline
        
        \textbf{STCFormer } & No&43.32 &  45.34 & 40.56 & \textcolor{red}{40.86} & 53.85 & 58.52 & \textcolor{red}{40.54} & \textcolor{blue}{40.04} & 61.91 &\textcolor{red}{ 60.36} & 48.59 & 49.50 & 48.41 & 32.50 & 34.73 & 46.60 \\ \hline
        
        \textbf{MHFormer } &No& {43.15} & 45.74 & \textcolor{blue}{40.08} & 40.99 & 53.09 & 52.41 & \textcolor{blue}{40.66} & 41.36 & 60.58 & 64.17 & 43.72 & 51.99 & 45.18 & 33.75 & 42.08 & 46.60 \\ \hline
        \textbf{STE} &No& 44.63 & 46.48 & 40.37 & 42.47 & 57.59 & 53.71 & 41.94 & 40.80 & 62.79 & 64.33 & 44.31 & 55.04 & 46.01 & 35.97 & 46.02 & 48.16 \\ \hline
        
        \textbf{PoseFormer } &No& 45.24 & 44.90 & 39.82 & 42.59 & \textcolor{blue}{48.36} & 52.52 & 42.14 & 42.09 & 57.00 & 60.72 & 46.09 & 52.39 & 47.43 & 31.85 & 40.07 & 46.2 \\ \hline
        \textbf{Att-3D } & No&47.33 & 49.18 & 42.11 & 44.62 & 54.74 & 55.55 & 42.66 & 42.07 & 60.21 & 65.99 & 46.40 & 55.77 & 46.79 & 31.62 & 39.86 & 48.3\\ \hline
        \textbf{VideoPose3D } &No& 49.59 & 47.53 & 44.37 & 47.86 & 57.03 & 57.99 & 45.70 & 44.72 & 59.38 & 66.81 & \textcolor{red}{38.39} & 49.61 & 51.03 & 36.94 & 34.11 & 49.4 \\ \hline
        \textbf{Uplift-Upsample } &Yes& 44.76 & \textcolor{blue}{42.82} & \textcolor{red}{39.49} & \textcolor{blue}{41.21} & 53.26 & \textcolor{red}{49.62} & 41.59 & 41.11 & \textcolor{blue}{56.99} & 61.30 & \textcolor{blue}{43.37} & 54.31 & \textcolor{red}{42.44} & \textcolor{red}{30.30}  & \textcolor{red}{29.35} & \textcolor{blue}{45.60} \\ \hline
         \textbf{PoseFormerV2 } & No&44.01 &{ 47.22} & 41.48 & 44.09 & 48.39 & 53.76&42.68 & 42.66 & 57.76 & 64.65 & 45.75 & 50.37&46.89&32.40&{ 34.22}&46.4\\ \hline
         \textbf{MHFormer } & Yes& \textcolor{blue}{41.25}&44.42&42.23&42.61&49.23&55.36&44.62&42.21&58.19&65.79&46.18&\textcolor{blue}{44.78}&45.75&31.20&33.68&45.83\\ \hline
        \textbf{P-STMO } &Yes& 44.01 & 46.19 & 40.52 & {41.28} & 55.94 & \textcolor{blue}{51.45} & 41.02 & \textcolor{red}{40.02} & 63.77 & 64.81 & 44.98 & 55.50 & 44.55 & 34.44 & 44.26 & 47.52 \\ \hline
        \textbf{T3D-CNN } &Yes& 48.63 & 50.17 & 49.69 & 50.86 & 55.13 & 63.88 & 47.76 & 47.21 & 63.11 & 71.14 & 52.32 & {49.13} & 55.97 & 39.11 & 40.96 & 52.3 \\ \hline
        \textbf{Ours (DTF)} & Yes&\textcolor{red}{37.20} & \textcolor{red}{41.23} & {40.29} & {41.47} & \textcolor{red}{47.43} & {52.41} & {41.30} & {40.73} & \textcolor{red}{53.40} & \textcolor{blue}{61.03} & {44.15} & \textcolor{red}{42.17} & \textcolor{blue}{44.10} & \textcolor{blue}{30.75} & \textcolor{blue}{30.19} & \textcolor{red}{43.19} \\ \hline
        \multicolumn{18}{|c|}{Protocol 2}\\
        \hline
          \textbf{Method} & \textbf{OAT}&\textbf{Dir.} & \textbf{Disc} & \textbf{Eat} & \textbf{Greet} & \textbf{Phone} & \textbf{Photo} & \textbf{Pose} & \textbf{Purch.} & \textbf{Sit} & \textbf{SitD.} & \textbf{Smoke} & \textbf{Wait} & \textbf{WalkD} & \textbf{Walk} & \textbf{WalkT.} & \textbf{Avg.} \\ \hline
           \textbf{PoseFormerV2 } & No&34.05 &36.86 & 33.80 & 35.88 & 36.94 & 41.17 & 33.18 & 32.73 & 46.05 & 51.92&37.45 & 36.88&36.23&{25.30}&27.35&36.4\\ \hline
        \textbf{STCFormer } & No&34.18 & 36.05 & 32.91 & \textcolor{red}{33.49} & 41.83 & 43.09 & \textcolor{red}{31.81} & \textcolor{blue}{30.93} & 49.57 & \textcolor{blue}{48.58} & 39.16 & 37.69 & 37.73 & 25.39 & 27.83 & 36.68 \\ \hline
        
        \textbf{MHFormer } & No&34.10 & 36.43 & 32.81 & 33.91 & 40.01 & 40.26 & \textcolor{blue}{31.90} & 32.31 & 48.13 & 51.17 & 36.33 & 38.43 & \textcolor{blue}{35.33} & 26.16 & 32.23 & 36.63 \\ \hline
        
        \textbf{STE } &No& 36.71 & 38.15 & 32.71 & 35.55 & 44.79 & 42.89 & 33.13 & 32.07 & 50.93 & 52.72 & 36.42 & 42.59 & 36.54 & 28.51 & 37.11 & 38.72 \\ \hline
        \textbf{PoseFormer } &No&34.79  &\textcolor{blue}{34.88} & 32.62 &34.67 & \textcolor{red}{36.26} & 40.12 & 32.22 & 32.03 & \textcolor{blue}{45.69} & \textcolor{red}{48.51} & 36.84 & {37.24} & 36.03 & 24.59 & 30.14 & 35.8 \\ \hline
        \textbf{Att-3D } & No&36.37 & 38.52 & 33.55 & 36.87 & 41.03 & 42.78 & 33.06 & 32.53 & 47.96 & 52.53 & 37.33& 40.66& 36.85 & {25.08} & 31.36 & 37.8 \\ \hline
        \textbf{VideoPose3D } &No& 37.16 & 36.86 & 35.16 & 38.76 &42.19  & 44.11 & 35.09 & 34.21 & 46.47 & 53.35 & 38.39 & 37.28 & 39.26 & 28.79 & {27.47} & 38.3 \\ \hline
         \textbf{Uplift-Upsample } & Yes&34.37 & \textcolor{red}{34.32} & \textcolor{red}{32.14} & 33.90 & 40.37 & \textcolor{red}{38.96} & 32.12 & 32.14 & 46.52 & 49.50 & \textcolor{red}{35.65} & 39.80 & 34.12 & 24.86 & \textcolor{red}{23.63} & \textcolor{blue}{36.14} \\ \hline
          \textbf{PoseFormerV2 } & No&34.05 &36.86 & 33.80 & 35.88 & 36.94 & 41.17 & 33.18 & 32.73 & 46.05 & 51.92&37.45 & 36.88&36.23&{25.30}&27.35&36.4\\ \hline
       
         \textbf{MHFormer } & Yes& \textcolor{blue}{33.22}&35.96&35.42&35.05&38.09&41.89&33.97&32.57&46.81&52.43&37.99&\textcolor{blue}{34.59}&35.87&\textcolor{blue}{24.83}&\textcolor{blue}{26.95}&36.38\\ \hline
        \textbf{P-STMO } & Yes&{34.01} & 36.97 & 32.98 & 34.05 & 39.66 & 40.25 & 32.60 & {31.61} & 49.09 & 50.63 & 36.40 & 38.74 & 35.35 & 26.30 & 31.01 & 36.64 \\ \hline
        \textbf{T3D-CNN } & Yes&36.44 & 38.83 & 39.15 & 40.72 & 41.72 & 47.35 & 36.7 & 35.55 & 49.44 & 55.7 & 41.38 & 36.42 & 43.45 & 29.32 & 33.22 & 40.4 \\ \hline
        \textbf{Ours (DTF)} & Yes&\textcolor{red}{30.01} & {34.92} & \textcolor{blue}{33.33} & \textcolor{blue}{33.89} & \textcolor{blue}{37.35} & \textcolor{blue}{39.48} & {32.08} & \textcolor{red}{30.80} & \textcolor{red}{42.81} & {49.33} & \textcolor{blue}{36.16} & \textcolor{red}{31.83} & \textcolor{red}{34.15} & \textcolor{red}{22.57} & {27.29} & \textcolor{red}{34.40} \\ \hline
    \end{tabular}
\end{table*}

\end{document}